%% file: root.tex
\documentclass[conference]{IEEEtran}
\usepackage{times}

\usepackage{multicol}
\usepackage[bookmarks=true]{hyperref}

\usepackage{amssymb}
\usepackage{todonotes}
\usepackage{booktabs}
\usepackage{multirow}
\usepackage{graphicx}
\usepackage{algorithm}
\usepackage{algorithmicx}
\usepackage{algpseudocode}
\usepackage{amsmath}
\usepackage{cleveref}
\usepackage{float}
\let\labelindent\relax
\usepackage{enumitem}
\usepackage{caption}
\usepackage{mathtools}
\usepackage{framed}

\newcommand{\multicomment}[1]{%
  \hfill\(\triangleright\) \begin{tabular}[t]{@{}l@{}}#1\end{tabular}%
}

\let\svthefootnote\thefootnote
\newcommand\freefootnote[1]{%
  \let\thefootnote\relax%
  \footnotetext{#1}%
  \let\thefootnote\svthefootnote%
}
\input{notation}

\begin{document}

\title{Show, Don't Tell: Detecting Novel Objects by Watching Human Videos}

\author{\authorblockN{
James Akl, 
Jose Nicolas Avendano Arbelaez,
James Barabas,
Jennifer L. Barry,
Kalie Ching,
Noam Eshed,\\
Jiahui Fu, 
Michel Hidalgo, 
Andrew Hoelscher,
Tushar Kusnur,
Andrew Messing,
Zachary Nagler,\\
Brian Okorn,
Mauro Passerino,
Tim J. Perkins,
Eric Rosen,
Ankit Shah,
Tanmay Shankar,
Scott Shaw}
\authorblockA{Robotics and AI Institute (RAI)\\
Cambridge, MA, USA\\}
}

\makeatletter
\patchcmd{\@maketitle}
  {\end{center}}  
  {%
    \end{center}  
    \vskip 1em
    {\centering
      \includegraphics[width=\textwidth]{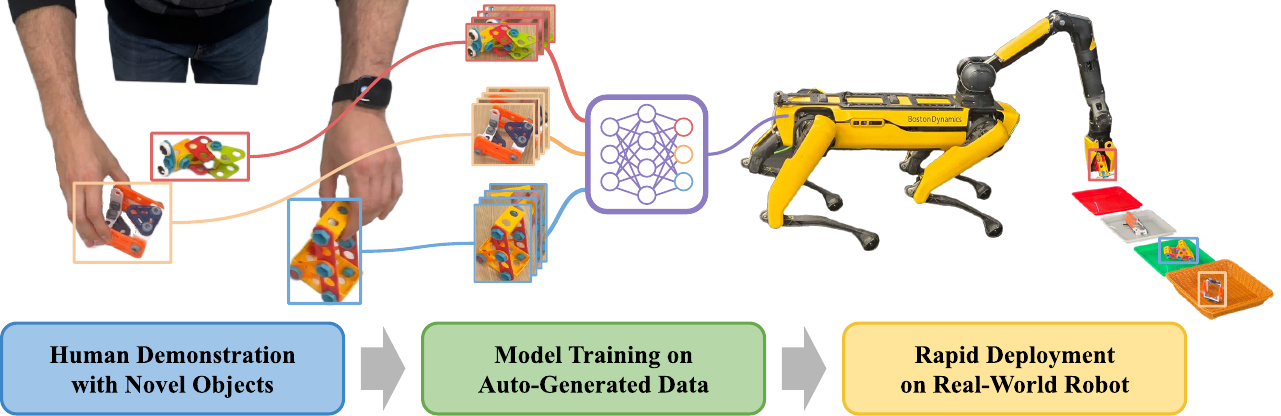}\par
      \captionof{figure}{Our system leverages human demonstrations and pretrained foundation models to automatically generate annotated training data for object detectors focused on the objects humans interact with. The resulting detectors can be deployed on robotic systems within minutes of capturing a demonstration and can distinguish novel objects that are difficult to describe using language alone, such as those found in assembly tasks.\label{fig:fig1}}\par
    }
    \vskip -0.5em
  }
  {}{}
\makeatother

\maketitle
\setcounter{figure}{1}  

\input{0-Abstract}

\IEEEpeerreviewmaketitle

\freefootnote{Authors are listed in alphabetical order. Robotics and AI Institute, Cambridge, MA, 02142. Correspondence to Brian Okorn \& Tanmay Shankar: \texttt{ \{bokorn, tshankar\}@rai-inst.com}.}
\input{1-Intro}

\input{2-Related}

\input{3-Method}

\input{4-Experiments}

\input{4a-On-Robot-Application}

\input{5-Conclusion}
\input{7-Acknowledgements}

\bibliographystyle{IEEEtran}
\bibliography{root}

\clearpage

\input{6-Appendix}


\end{document}

%% file: notation.tex
\usepackage{amsmath}

\newcommand{\eg}[0]{{\em e.g.,~}}

%% file: 0-Abstract.tex
\begin{abstract}
How can a robot quickly identify and recognize new objects shown to it during a human demonstration? Existing closed-set object detectors frequently fail at this because the objects are out-of-distribution. While open-set detectors (\eg VLMs) sometimes succeed, they often require expensive and tedious human-in-the-loop prompt engineering to uniquely recognize novel object instances. 
In this paper, we present a self-supervised system that eliminates the need for tedious language descriptions and expensive prompt engineering by training a bespoke object detector on an automatically created dataset, supervised by the human demonstration itself. 
In our approach, ``Show, Don't Tell," we \textit{show} the detector the specific objects of interest during the demonstration, rather than \textit{telling} the detector about these objects via complex language descriptions. By bypassing language altogether, this paradigm 
enables us to quickly train bespoke detectors tailored to the relevant objects observed in human task demonstrations. We develop an integrated on-robot system to deploy our ``Show, Don't Tell" paradigm of automatic dataset creation and novel object-detection on a real-world robot. Empirical results demonstrate that our pipeline significantly outperforms state-of-the-art detection and recognition methods for manipulated objects, leading to improved task completion for the robot.
\end{abstract}

%% file: 1-Intro.tex
   
\section{Introduction}
\label{sec:intro}
Consider a robot learning to perform a task by watching a person demonstrate the task, akin to an on-the-job training setting. The robot must recognize what objects the person interacts with and how these objects move in the environment. For example, when a person demonstrates sorting components into a kit, the robot must be capable of identifying and differentiating relevant components.

Large-scale object recognition models are the prevalent method for detecting objects. Closed-set object detectors~\cite{li2021grounded,radford2021learning} perform well on common object categories \cite{lin2015microsoftcococommonobjects} but are not intended to detect objects outside of their training distribution, such as custom manufactured parts or assembled objects.  Open-set detectors~\cite{Cheng2024YOLOWorld,jiang2025rexomni,liu2023groundingdino} can identify some novel objects beyond their training support, but they often struggle with unique instance recognition~\cite{geigle2024does}—including parts~\cite{zeng2024compchall}—or need complex prompt engineering~\cite{chen2025unleashing} so tedious that the community has turned to automated prompt tuning~\cite{du2022learning}.

How can we bypass the need for complex language descriptions to prompt these Vision-Language Models (VLMs)?  We introduce the “Show, Don’t Tell” paradigm, shown in Figure~\ref{fig:fig1}, where the core idea is to \textit{show} the detector objects of importance, rather than \textit{tell} the detector of these objects via complex language descriptions.  We instantiate this paradigm in the three key contributions of this paper:
\begin{enumerate}
    \item A \textit{Salient Objects Dataset Creation} pipeline (SODC) that uses off-the-shelf models to automatically create labeled datasets of manipulated objects from a single human demonstration.
    \item A \textit{Manipulated Objects Detector} (MOD) that uses the dataset to train a lightweight novel-object detector that is more accurate and faster than large open-set models for task-relevant objects.
    \item A fully integrated on-robot system that, given a video of a single human demonstration, runs the dataset creation, trains a manipulated objects detector, and infers the sequence of picks and places in order to replicate how a human sorts previously unseen objects.
\end{enumerate}



%% file: 2-Related.tex
\section{Related Works}
\label{sec:related}


Our proposed system is aimed at fast adaptation to short human interactions with novel, hard-to-name objects. We review various object detection and recognition works below, along with a short review of modeling human-object interactions and learning from demonstration.  We focus on how these systems deal with novel objects and the ways we could integrate the dataset generated by SODC.

\subsection{Object Detection \& Recognition} 
\label{sec:rw:obj}
Object detection and recognition have been long standing problems in the computer vision \cite{SALARI2022129,zou2023objectdet} and robotics communities \cite{bai2020objectdetectionrobotics} that have served as foundational tasks in machine perception.  There are currently two main approaches to object detection: closed-set detection and vision-language models.


\subsubsection{Closed Set Detectors}
Closed set object detectors use neural-net architectures trained with large scale (closed-set) datasets such as COCO \cite{lin2014microsoft, lin2015microsoftcococommonobjects}. Early  approaches relied on two-stage proposal-classification pipelines~\cite{girshick2014rich, redmon2016yolo, ren2015faster}. These were later replaced with transformer-based detectors \cite{carion2020end, zhang2022dino, rf-detr} that use query vectors to represent objects. Despite efforts to broaden the scope of these detectors \cite{zhou2022detecting}, these approaches collectively focused on closed-set detection, with detecting new or novel objects requiring fine-tuning these models to the target domain.  Our approach enables automatic fine-tuning by observing a single human demonstration.

\subsubsection{Vision-Language Models for Object Detection}
More recently, the community has explored leveraging large-scale pre-training to include the task of object detection within the capabilities of foundation models. 
One way to accomplish this has been to extend closed-set detectors beyond their training support with the incorporation of language into the model, either by contrastive training to co-embed textual queries with images~\cite{li2021grounded,radford2021learning}, using multimodal LLMs to directly recognize objects in images~\cite{hurst2024gpt, team2023gemini, team2025gemini}, or training LLMs to output pointing locations on an input image to serve as a proxy for the detection task~\cite{deitke2024molmo}. 
We employ three VLMs as baseline approaches for object detection: YoloWorld~\cite{Cheng2024YOLOWorld}, a lightweight model that co-embeds language and visual features; GroundingDINO~\cite{liu2023groundingdino}, which conducts grounded pre-training on a transformer based detector; and RexOmni~\cite{jiang2025rexomni}, a large multi-task multi-modal LLM trained for detecting arbitrary objects.  We direct the reader to \cite{feng2025_VLMDetection} for a comprehensive review of using VLMs for object detection.

\textit{The need for adaptation}: While the closed-set approaches can be trained to high accuracy on the training set, a single checkpoint is not expected to succeed at detecting a novel object. In contrast, although VLMs do not have an explicitly defined closed-set of objects, they struggled with detecting novel objects in our evaluations (Section~\ref{sec:experiments}), primarily due to the difficulty of naming novel objects.  For example, VLMs failed to recognize the top object in Figure~\ref{fig:fig1} with prompts like ``green yellow red worm toy" or ``eyes on red green yellow toy."  

\textit{Multi-modal prompting}: One method of avoiding the complexity of language prompting is \textit{visual} or \textit{multimodal} prompting where the VLM is provided with a picture of the object to detect along with an imperfect textual prompt. For example, MQ-Det \cite{xu2023multimodal} re-weights the tokens of the text prompt conditioned on the exemplar image, while  Wu et al. \cite{wu2025visual} projects image embeddings into the same space as text embeddings for models such as GLIP \cite{li2021grounded}. However, in the case of novel objects, we must answer the question ``where do the visual prompts (e.g., pictures of the objects) come from?" The Salient Objects Dataset Creation pipeline in our work serves to answer to this question; and as a method for automatic visual or multi-modal prompting.

\textit{The Show, Don't Tell paradigm}: 
Other works \cite{gupta-etal-2022-show, chen2026showdonttellmorphing, komp2024showdonttelllearning} have referred to the ``Show, Don't Tell" phrase previously, but do so in widely different domains, and typically use the phrase to describe the use of arbitrary (non-visual) exemplars to instantiate a concept over using language input alone. 
In contrast, we introduce the ``Show, Don't Tell" paradigm in the context of relying on \textit{visual} prompts as instance level supervision over pure language descriptions.

\label{sec:sodc}
\begin{figure*}[t!]
    \centering
    \includegraphics[width=\textwidth]{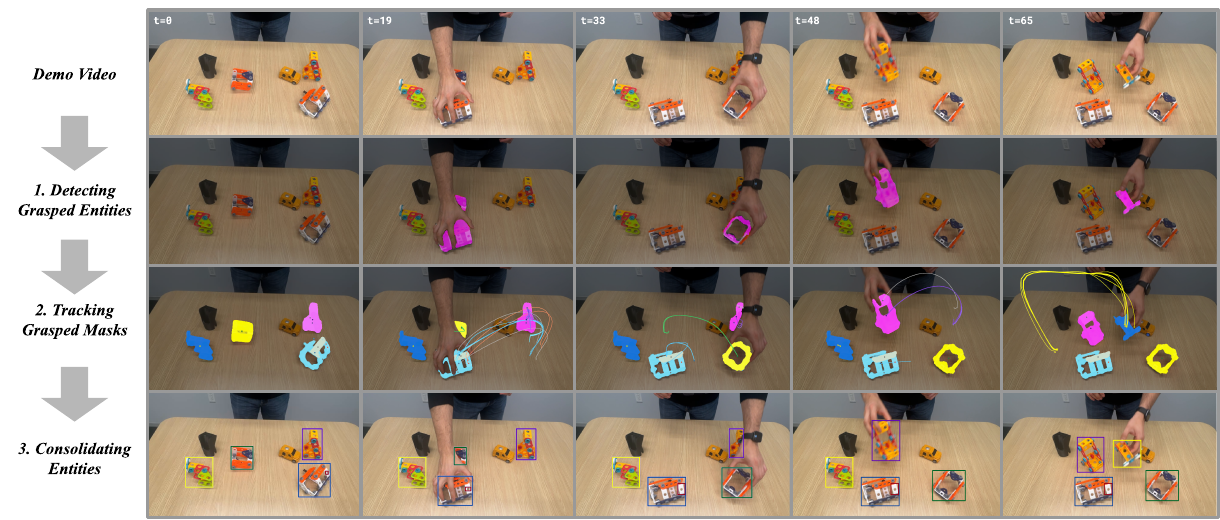}
    \caption{Our Salient Objects Dataset Creation (SODC) pipeline: \textit{1. Detecting Grasped Entities:} A Human-Object-Interaction detector is used to detect and segment grasped objects in each frame (shown in pink). \textit{2. Tracking Grasped Masks:} Entities are tracked over time. Note that each grasp segmentation is used as a seed to a tracking algorithm, resulting in multiple tracks per grasped object, with each track represented as a colored trajectory. \textit{3. Consolidating:} Tracks are clustered across space and time to identify individual objects.}
    \label{fig:sodc}
\end{figure*}

\subsection{Modeling Human-Object Interactions}
\label{sec:rw:human}
For this work, we define task-relevant objects as those a human manipulates, which is broadly situated in the context of modeling hand-object interactions and co-training for object and activity detection. Prior research into this problem~\cite{Darjana2025_EgoSurgeryHTS, liu2014recognizing, Narasimhaswamy2024_Hoistformer, Pei2025_HOD, Shan2020_100DOH} typically models objects \textit{while in-hand}, i.e., while they are being interacted with.  A key distinction here is we learn to detect objects using in-hand interactions as a supervision tool to train detectors for in-the-wild, out-of-hand detections. While our manipulated objects detector is intended to recognize objects in any context, we rely on in-hand object detection and segmentation to seed our training data. Specifically, we chose HOIST-Former \cite{Narasimhaswamy2024_Hoistformer} as our in-hand object detector. However, our approach is compatible with any in-hand object detector that returns a pixel-level mask. 

\subsection{Learning from Demonstration}
While our final on-robot deployment is a version of learning from demonstration~\cite{ravichandar2020recent} in that a robot copies a series of picks and places from a human, our focus is on using the human demonstration for the novel object detection problem.  Behavior-cloning or imitation learning approaches~\cite{team2025gemini, intelligence2025pi_, intelligence2025pi6, vosylius2024instant, zhong2025survey} fold the object detection task into end-to-end models but also require on-robot training data, which our approach does not.  Many other learning from demonstration approaches require object detections or bounding boxes as part of their training or input~\cite{jiang2023vima, perez2017c, mueller2018robust, hayes2014discovering, toris2015unsupervised, fitzgerald2015visual} and could thus use our approach in their pipelines.

%% file: 3-Method.tex
\section{Approach}
\label{sec:methods}

\begin{figure*}[t]
    \centering
    \includegraphics[width=\textwidth]{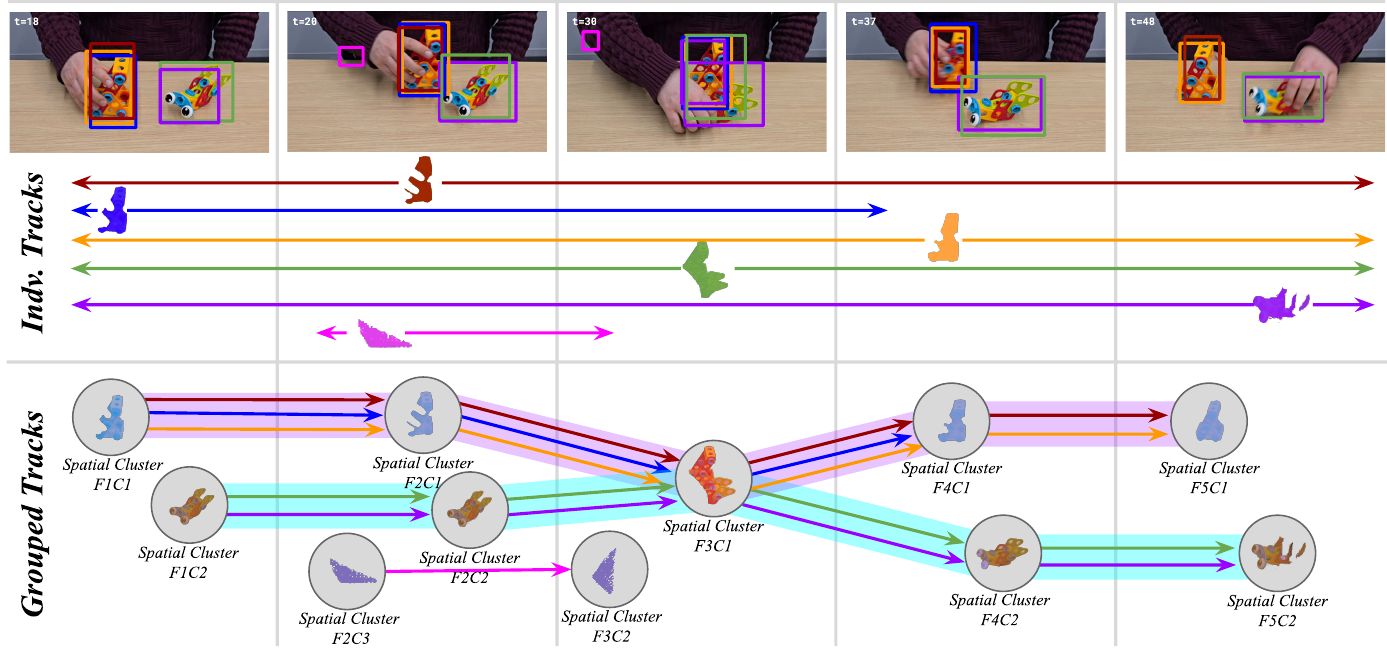}
    \caption{Track Clustering: Combining multiple bounding box tracks (``Indv. Tracks" with colors corresponding to their bounding boxes) into per-object tracks (highlighted light purple and cyan). First, the bounding boxes in each frame are spatially clustered (F1C1, F1C2, etc.). Then the tracks are temporally grouped if they traverse the same or highly similar sequences of spatial clusters. Note that although five bounding boxes are clustered spatially at t=30, they are grouped into two final temporal groups because those bounding boxes are not clustered together in all frames.  The short magenta track is discarded as noise because its temporal group does not include enough tracks.}
    \label{fig:clustering}
\end{figure*}

\input{3a-Problem-Formulation}

\subsection{Salient Objects Dataset Creation}

We first introduce the Salient Objects Dataset Creation pipeline (SODC), which, from human interactions with novel objects, creates a dataset of images with bounding boxes around all objects a human has manipulated at any point in the video.  Example images from such a dataset are shown in Figure~\ref{fig:sodc}-(3).  The pipeline has three steps:
\begin{enumerate}[wide = 5pt]
    \item Detect entities manipulated by humans in a video
    \item Track these entities across the video
    \item Cluster these tracked entities into a labeled dataset
\end{enumerate}

The full pipeline is shown in Figure~\ref{fig:sodc} and we go through each step in more detail below.

\subsubsection{Detecting Grasped Entitites}
\label{sec:hoistformer}
Given a video of a human interaction with novel objects, we first identify the objects the human manipulates.  We use HOIST-Former for its accuracy and ability to detect arbitrary objects~\cite{Narasimhaswamy2024_Hoistformer}, though other approaches such as~\cite{Pei2025_HOD, Shan2020_100DOH} or even prompt-conditioned segmentation~\cite{carion2025sam} could be employed as well. Given an input set of images, HOIST-Former outputs a list of segmentation masks for every frame in which the human is grasping an object (step 1 in Figure~\ref{fig:sodc}).  Note that, while HOIST-Former exhibits some some label persistence over time, we found it too noisy for practical use. As a result, our system considers each frame independently - there is no association between the grasped mask in two different frames even if the human was grasping the same object in both of those frames.

\subsubsection{Tracking Grasped Masks}
\label{sec:tracking}
The output of HOIST-Former from Section~\ref{sec:hoistformer} is a set of segmentation masks only for frames in the video in which an object is being manipulated. By itself, this is insufficient data from which to create an object detector and largely contains frames in which the object is heavily occluded by the human hand. We therefore track objects across the entire video, including the frames in which they're not manipulated, by ``seeding" a tracking algorithm with the masks returned by HOIST-Former and tracking forwards and backwards from that frame.  Any keypoint~\cite{doersch2023tapir,zhang2025tapip3d} or mask tracking algorithm~\cite{carion2025sam,ravi2024sam, ravi2024samurai} can be used for the tracking; in our pipeline we chose SAMURAI~\cite{ravi2024samurai}. As shown in step 2 of Figure~\ref{fig:sodc}, this provides us with a list of ``tracks'' of masks, one per frame in which HOIST-Former detected something grasped.  Note that we have far more tracks than objects as HOIST-Former likely detected each object in multiple frames.

\begin{figure*}[t]
    \centering
    \includegraphics[width=\textwidth]{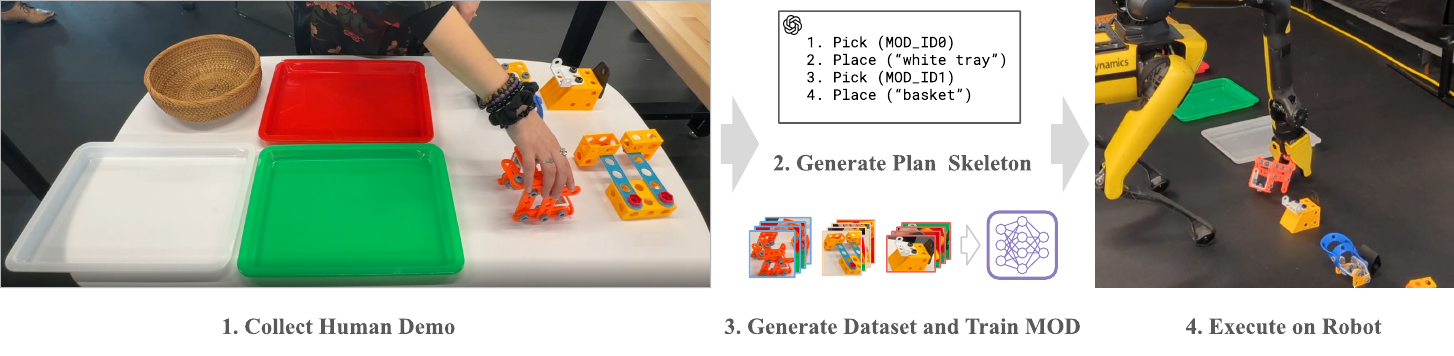}
    \caption{The on-robot application flow: (1) The participants make novel objects and demonstrate a sort to the camera.  From the video the robot generates (2) a plan skeleton, (3) a salient objects dataset, and a manipulated objects detector.  (4) The robot executes the sort.  Note that the basket the robot uses is not identical to the basket used in the human demonstration since we use a VLM to recognize objects the human does not manipulate directly.}
    \label{fig:application}
\end{figure*}

\subsubsection{Consolidating Across Space and Time}
\label{sec:clustering}

At the end of the tracking described in Section~\ref{sec:tracking}, we have a large set of ``tracks" of various masks, many of which represent the same object.  The last step in our algorithm is to cluster together the tracks representing a single object.  To do this we first convert masks to bounding boxes.  Intuitively, at this point, bounding boxes that significantly overlap each other \textit{in all frames} represent the same object.  For example in Figure~\ref{fig:clustering}, the red, blue, and yellow bounding boxes travel together throughout almost all the frames, as do the green and purple boxes.  In other words, the red, blue, and yellow tracks ``look" the same over time, as do the green and purple tracks.

To identify the tracks that ``look" the same, we cluster in two dimensions: space and time.  We start by clustering spatially, using DBSCAN~\cite{ester1996density} with an IoU distance function to cluster overlapping bounding boxes in each frame.  The result of this is that each bounding box is assigned a cluster per-frame.  For example, at t=18 (first frame) in Figure~\ref{fig:clustering}, the red, yellow, and blue bounding boxes are all assigned to cluster F1C1 while the purple and green are assigned to F1C2.  For each bounding box track, we create a ``cluster track" - the list of all clusters across frames to which the corresponding bounding box was assigned.  For example the red track in Figure~\ref{fig:clustering} has the cluster track [F1C1, F2C1, F3C1, F4C1, F5C1].  Bounding box tracks representing the same object should have identical or very similar cluster tracks - e.g. in Figure~\ref{fig:clustering}, red and yellow both have the cluster track [F1C1, F2C1, F3C1, F4C1, F5C1] and blue has [F1C1, F2C1, F3C1, F4C1] while purple and green both have the track [F1C2, F2C2, F3C1, F4C2, F5C2].  Thus we cluster temporally by creating the cluster track for each bounding box track, computing the Jaccard score~\cite{jaccard1901distribution} between each pair of cluster tracks, and clustering bounding box tracks with sufficiently similar scores.  We remove any clusters that have an insufficient number of bounding box tracks in them.

Note that the combination of temporal and spatial clustering makes us robust to noise and temporary object occlusions.  At t=30 (third frame) in Figure~\ref{fig:clustering} the red, blue, yellow, green, and purple clusters are all in the same place.  Pure spatial clustering would conclude they all represent the same object.  The temporal clustering shows us that they are, in fact, two different objects.  At t=20 and t=30 (second and third frames) in Figure~\ref{fig:clustering}, we have a brief noisy bounding box track.  However, it does not have a sufficient number of tracks in its final cluster to be included.

The final output of the clustering is a set of images with labeled bounding boxes with one label per different manipulated object in the original human video.

\subsection{Manipulated Objects Detector}
\label{sec:just_mod}
We fully instantiate the ``Show, Don't Tell" paradigm introduced in Section~\ref{sec:intro} by using the dataset created by SODC to train a task-specific object detector on all objects a human touches in a demonstration.
SODC affords us a dataset of images, bounding boxes, and associated labels for each of the manipulated objects in the demonstration. Note that unlike the vanilla HOIST-Former output, this dataset contains images with the objects both in and out of the human's hands.
We train a lightweight object detector on this dataset by fine-tuning a pretrained F-RCNN model \cite{FRCNN} with a ResNet50 backbone, minimizing standard RCNN losses (classification and objectness) \cite{FRCNN}. Training takes $\sim$3-4 minutes on 4 T4 GPUs. We apply augmentations to the model inputs, such as random flipping, distortion, brightness, contrast, color, crop, zoom, blur, and affine transformations.  These augmentations are detailed in Appendix~\ref{app:mod}.

\subsection{On Robot Application}
\label{sec:application}

We showcase our manipulated objects detector with an end-to-end robotics application that simulates a sorting or kitting task and can be shared with the general public. In this task:
\begin{enumerate}[wide = 5pt]
    \item We collect a video of a human participant sorting a set of novel objects.
    \item The robot processes the video, creating a salient objects dataset, training a manipulated objects detector, and generating a \textit{plan skeleton} of the sequence of picks and places demonstrated by the human.
    \item The robot copies the demonstrated sort of these novel objects in a different environment.
\end{enumerate}
We present an overview of the application flow in Figure~\ref{fig:application}.

We generate the plan skeleton by prompting a large language model (ChatGPT-4o~\cite{openai2024gpt4technicalreport}) with the video and a request to break it into picks and places with names for each of the place objects.  We use IDs from the manipulated objects detector for pick objects.  For example, for a single pick and place, the plan skeleton might be \texttt{[Pick(MOD\_ID0), Place("basket")]}, which tells the robot to pick the object with ID 0 from the manipulated objects detector and place it in a basket. Full details of plan skeleton generation can be found in Appendix~\ref{app:plan_skeleton}.  

At execution time, the robot first uses the manipulated objects detector to find all of the novel objects the participants built. Place objects (which the human did not directly manipulate) are found using a vision language model~\cite{liu2023groundingdino}. Note that this allows place objects to be semantic - e.g., the robot can use a differently shaped ``basket" than the person did. As objects are found, they are stored in a scene graph, allowing us to aggregate point clouds over time.  More details of the online perception system can be found in Appendix~\ref{app:scenegraph}.  Once the robot has found all of the necessary objects, it carries out the picks and places in the order specified by the plan skeleton.  More details on the robot's execution can be found in Appendix~\ref{app:impl}.  Further details of the deployment can be found in Section~\ref{sec:exp:application}.

The entire understanding pipeline, including plan skeleton generation and the manipulated objects detector training, takes approximately 4-7 minutes on a 15 second video. 


%% file: 3a-Problem-Formulation.tex
\subsection{Problem Formulation and Assumptions}
Our problem is detecting \textit{novel objects} in \textit{any context} from a \textit{single human interaction}.  \textit{Novel objects} are objects the system has never seen before even in training data like those shown in Figure~\ref{fig:fig1}.  The input to the system is a single RGB video of a human interacting with novel objects.  The output is 1) a set of images with bounding boxes around manipulated objects and 2) a detector that can detect these novel objects in arbitrary images.  For this work we make the following assumptions:
\begin{itemize}
    \item Any object with which the human interacts for a significant time is task-relevant
    \item Any object with which the human does not interact is not task-relevant.
\end{itemize}


%% file: 4-Experiments.tex
\section{Experiments}
\label{sec:experiments}
Our experiments serve to answer three questions: Firstly, how well does our proposed MOD (and therefore, our full ``Show, Don't Tell" paradigm) perform on novel-object detection in comparison with state-of-the-art object-detection methods, such as VLMs?  Secondly, is the automatic data generation useful?   And thirdly, can our proposed approach enable a real robot to perform a task requiring detecting novel objects?

\input{4-1-table1}

\subsection{Experimental Setup}
\label{sec:exp:setup}
To answer the first question, we carry out perceptual evaluations of the MOD pipeline and a set of VLMs on classical object detection tasks across a variety of datasets.
\subsubsection{Datasets}
We use the following datasets. Since our work operates in the \textit{novel} object detection setting, we do not evaluate on common closed-set object detection datasets such as COCO \cite{lin2015microsoftcococommonobjects}. 
\begin{itemize}[wide = 0pt]
    \item \textit{Meccano Dataset:}
        The Meccano dataset \cite{ragusa_MECCANO_2023} includes 20 egocentric RGB videos of a subject constructing a toy motorbike along with object bounding boxes and action labels per frame. 
        The videos in the dataset range from 12 to 34 minutes in length, but 19 of them follow the same construction process, described step-by-step in an assembly manual.  We use those 19 so that we can use separate videos in training and testing.
        We extract 20 second snippets at 10 percent increments through the duration of each video and divide these into train and test sets per increment.
    \item \textit{In House Human \& Robot Video Dataset (In-House Dataset 1):}
        We collected an in-house dataset consisting of 54 human demonstrations 
        performing with different types of interactions (single-handed, bimanual, interacting with subsets of objects, etc.), and the object sorting task described in \cref{sec:exp:application} with different object sets and backgrounds, serving as our train set. The dataset also contains 18 videos from a Spot robot hand camera dynamically viewing subsets of these novel objects, annotated with object bounding boxes and labels, serving as our test set. This dataset includes 12 novel objects. 
    \item \textit{In House Human Video Dataset (In-House Dataset 2):}
        We collect an additional in-house human dataset of 61 videos including interactions for both the sorting task described in \cref{sec:exp:application} and freeform movement of the objects. This dataset includes 17 novel objects.

More details of the in-house datasets can be found in Appendix~\ref{app:objects} 

\end{itemize}

\subsubsection{Baselines}
We compare our proposed Manipulated Objects Detector (MOD) against a span of state-of-the-art object detection approaches. 
\begin{itemize}[wide = 0pt]
    \item RexOmni \cite{jiang2025rexomni} -- A state-of-the-art multimodal LLM designed for object detection. 
    \item GroundingDINO \cite{liu2023groundingdino} -- A transformer based detector (DINO \cite{zhang2022dino}) with grounded pre-training. 
    \item YoloWorld \cite{Cheng2024YOLOWorld} -- A lightweight VLM based detector. 
\end{itemize}
Since our focus is novel-object detection, we do not compare against off the shelf closed set detectors, as they are not intended to detect these novel objects without further training.

\subsubsection{Prompt Generation}
\label{sec:exp:promptgen}
Each of our VLM baselines require language prompts to detect the queried objects. We carry out comparisons against the following prompting strategies - 
\begin{itemize}[wide = 0pt]
    \item GPT based prompting -- We use SODC to generate target object patches and then generate a prompt with ChatGPT-4o~\cite{openai2024gpt4technicalreport}.  This is the closest comparison to MOD as, like MOD, it is an automated process that uses the video.
    \item Human Prompt -- A human user comes up with a single prompt to detect each object in training videos, validated by feeding these prompts to GroundingDINO~\cite{liu2023groundingdino}. We ensure these are reasonable linguistic descriptions of the objects alone (e.g., a ``red construction toy"). We share these prompts across VLM baselines.  This requires significant human effort (see Section~\ref{sec:human_prompting}) to generate the prompts, which are validated ahead of time, and therefore represents close to a ``best case" scenario for VLM object detection. 
\end{itemize}

\subsubsection{Metrics}
We conduct evaluations on classical object detection tasks using mean average precision (mAP) reported across IoU thresholds (0.5-0.95); mean average recall (mAR) for one detection; and F1 score, Precision, and Recall averages across IoU thresholds (0.5-0.95) as is standard~\cite{Cheng2024YOLOWorld, jiang2025rexomni}. 

\subsubsection{Evaluation Pipeline}
Our proposed MOD is trained on a single human video. To evaluate it, for each dataset, we train a single MOD on a single video (video snippet for the Meccano Dataset), and evaluate it on all the corresponding test set videos. The baseline approaches are each evaluated on all videos in the test set of each dataset.

\subsection{Quantitative Evaluations}

We document the results of our perceptual evaluations in Table~\ref{table:1}, and present additional metrics in Appendix~\ref{app:exp}. On each of our In-House datasets, our proposed MOD achieves significantly higher mean average precision (mAP) \textit{and} mean average recall (mAR) than all other baselines. 
We observe this trend across different scales of baseline models used (from large models such as RexOmni \cite{jiang2025rexomni}, to similar model sizes, as in YoloWorld \cite{Cheng2024YOLOWorld}), as well as types of prompts that are provided to those VLMs (both automatically generated from GPT and human engineered prompts). 

The overall performance of our proposed MOD is good enough to deploy on robot (Section~\ref{sec:exp:application}) despite having seemingly low absolute values of the mAP.
 The objects in our In-House dataset serve as a good representation of compositionally novel objects commonly encountered in assembly or factory settings and are not easily described by natural language. 
The VLM baselines either fail to detect these novel-objects entirely (YoloWorld \cite{Cheng2024YOLOWorld}) or succeed at category level detection but struggle at instance disambiguation and recognition for these novel objects, as in the case of RexOmni \cite{jiang2025rexomni} and GroundingDINO~\cite{liu2023groundingdino}. 
In contrast, our MOD approach is able to successfully perform instance level recognition and disambiguate between similar objects, leading to better metric performance than the VLM counterparts. 

On the Meccano dataset~\cite{ragusa_MECCANO_2023}, we observe that MOD again outperforms every baseline on reported mAP and mAR, with the exception of GroundingDINO~\cite{liu2023groundingdino} using human prompts. This is likely because the human prompts used are validated with GroundingDINO in the loop, thereby biasing evaluations in favor of GroundingDINO. GroundingDINO fails to effectively recognize objects when these tuned human prompts are replaced with (otherwise reasonable) GPT generated prompts -- further evidence of the brittleness of the VLM based approach to detection, motivating our ``Show, Don't Tell" approach. 
Outside of GroundingDINO, MOD is the most performant model and achieves the best precision in recognizing objects. 


\subsection{Inference Time}
MOD and YoloWorld~\cite{Cheng2024YOLOWorld} each take $\sim$100ms to run inference on an image. GroundingDINO \cite{liu2023groundingdino} takes $\sim$400ms, and RexOmni \cite{jiang2025rexomni} takes $\sim$1-2s. Our proposed MOD is quantitatively more performant than VLM baselines, and it is just as fast as the quickest of these -- YoloWorld. This is particularly helpful in running our detector online on-robot. 

\subsection{Human Prompting}
\label{sec:human_prompting}

\begin{figure}
    \centering
    \includegraphics[width=\linewidth]{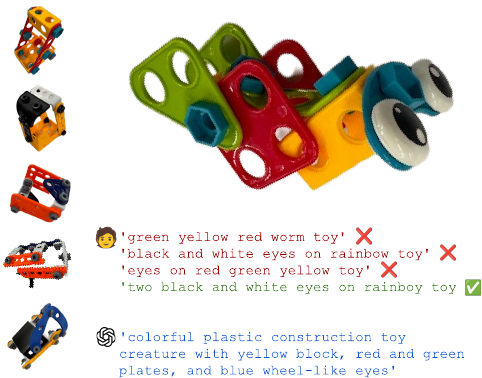}
    \caption{Examples of objects constructed by human participants.  Red text are human generated prompts from which VLMs could not detect the large object while green are successful prompts, generated iteratively in a single labeling session.  The blue text is the prompt generated by Chat GPT-4o.}
    \label{fig:objects}
\end{figure}

\begin{figure*}[t]
    \includegraphics[width=\textwidth]{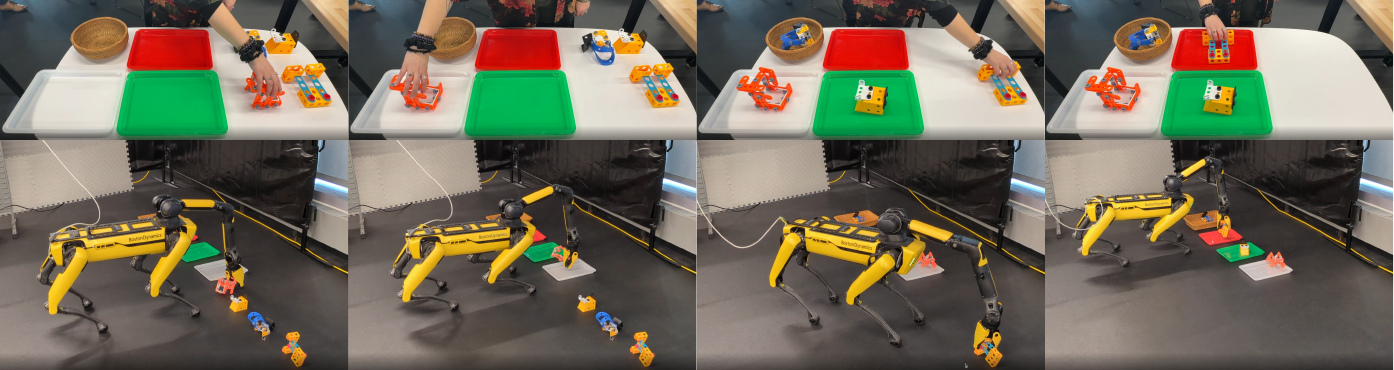}
    \caption{A timelapse of the on-robot application showing the human demonstration and the robot's execution.  The full demonstration and execution had four picks+places.}
    \label{fig:app_timelapse}
\end{figure*}

To answer question 2 ("Is the SODC pipeline useful?") we looked into the difficulty of creating the human prompts we used for the VLMs in Section~\ref{sec:exp:promptgen}.  We gave 16 human participants five tries to find prompts that GroundingDINO~\cite{liu2023groundingdino} or Detic~\cite{zhou2022detecting} could use to successfully detect the six novel objects made from construction kits shown in Figure~\ref{fig:objects} in scenes with multiple objects.  More details on this experimental setup are given in Appendix~\ref{app:prompt_tool}.  Quantitatively, on average it took humans over three attempts to create a usable prompt, and 43\% of the time humans failed to generate a workable prompt within the allotted five tries.  Qualitatively, we saw that creative visual concept prompting (e.g., referring to objects as sled, claw, shoe, etc.) was more effective than simply describing the objects (e.g., ``multi-part construction toy with gray, white, and yellow parts”).  Thus we conclude that automatically generating a labeled dataset that can be used instead of prompting (whether we do that by fine-tuning, visually prompting, or, as done here, training a bespoke detector) saves significant human effort.

%% file: 4-1-table1.tex
{\renewcommand{\arraystretch}{1.3}
\begin{table*}[]
\begin{center}
\caption{Quantative evaluation of object detection approaches on standard detection metrics. Higher is better.  Highest values are bolded, highest values not requiring human prompting are underlined.}
\label{table:1}
\resizebox{\linewidth}{!}{
\begin{tabular}{@{}cccccccc@{}}
\toprule
\multirow{2}{*}{Datasets} & \multicolumn{2}{c}{Approach} & \multicolumn{5}{c}{Metrics} \\ \cmidrule(l){2-8} 
 & Model & Prompt & mAP\textsubscript{0.5-0.95} & mAR\textsubscript{1} & F1\textsubscript{\rm 0.5-0.95} & Precision\textsubscript{\rm0.5-0.95} & Recall\textsubscript{0.5-0.95} \\ \midrule

\multirow{7}{*}{\begin{tabular}[c]{@{}c@{}}Meccano  \\ Dataset \cite{ragusa_MECCANO_2023} \end{tabular}}
& \multirow{2}{*}{RexOmni \cite{jiang2025rexomni}} & GPT-Prompt & 0.00 & 0.00 & 0.13 & 0.31 & 0.10 \\
&  & Human-Prompt & 0.05 & 0.09 & \textbf{0.30} & 0.59 & \textbf{0.23} \\
 & \multirow{2}{*}{GroundingDINO \cite{liu2023groundingdino}} & GPT-Prompt & 0.00 & 0.00 & \underline{0.22} & 0.41 & \underline{0.17} \\
&  & Human-Prompt & \textbf{0.19} & \textbf{0.26} & 0.24 & 0.46 & 0.18 \\
& \multirow{2}{*}{YoloWorld \cite{Cheng2024YOLOWorld}} & GPT-Prompt & 0.00 & 0.00 & 0.00 & 0.01 & 0.00 \\
 &  & Human-Prompt & 0.03 & 0.03 & 0.00 & 0.01 & 0.00 \\
 & MOD \textbf{(Ours)} & -- & \underline{0.06} & \underline{0.10} & 0.18 & \underline{\textbf{0.71}} & 0.12 \\ \midrule

\multirow{7}{*}{\begin{tabular}[c]{@{}c@{}}In-House \\  Dataset \#1 \end{tabular}} & \multirow{2}{*}{RexOmni \cite{jiang2025rexomni}} 
  & GPT-Prompt & 0.06 & 0.09 & \underline{\textbf{0.98}} & \underline{\textbf{1.00}} & \underline{\textbf{0.97}} \\ &
  & Human-Prompt & 0.04 & 0.06 & 0.96 & \textbf{1.00} & 0.93 \\
& \multirow{2}{*}{GroundingDINO \cite{liu2023groundingdino}} 
  & GPT-Prompt & 0.00 & 0.01 & 0.73 & 0.94 & 0.69 \\ & 
  & Human-Prompt & 0.04 & 0.08  & 0.87 & \textbf{1.00} & 0.82 \\
& \multirow{2}{*}{YoloWorld \cite{Cheng2024YOLOWorld}} 
  & GPT-Prompt & 0.00 & 0.00 & 0.30 & 0.51 & 0.27 \\
 &  & Human-Prompt & 0.02 & 0.03 & 0.45 & 0.61 & 0.39\\
 & MOD \textbf{(Ours)} & -- & \underline{\textbf{0.10}} & \underline{\textbf{0.17}} & \underline{0.92} & \underline{\textbf{1.00}} & 0.87 \\ \midrule
 
\multirow{4}{*}{\begin{tabular}[c]{@{}c@{}}In-House \\  Dataset \#2 \end{tabular}} & \multirow{1}{*}{RexOmni \cite{jiang2025rexomni}} 
 & GPT-Prompt & 0.09 & 0.12 & \underline{\textbf{0.99}} & \underline{\textbf{1.00}} & \underline{\textbf{0.99}} \\
  & \multirow{1}{*}{GroundingDINO \cite{liu2023groundingdino}} 
 & GPT-Prompt & 0.08 & 0.10 & 0.98 & \underline{\textbf{1.00}} & 0.96 \\
 & \multirow{1}{*}{YoloWorld \cite{Cheng2024YOLOWorld}} 
 & GPT-Prompt & 0.02 & 0.03 & 0.82 & \underline{\textbf{1.00}} & 0.77 \\
  & MOD \textbf{(Ours)} & -- & \underline{\textbf{0.15}} & \underline{\textbf{0.19}} & 0.95 & \underline{\textbf{1.00}} & 0.91 \\ 

 \bottomrule

\end{tabular}
}
\end{center}

\end{table*}
}

%% file: 4a-On-Robot-Application.tex
\subsection{On-Robot Application}
\label{sec:exp:application}

To answer our final question (``Can our proposed approach enable a real robot to perform a task requiring detecting novel objects?"), we deployed the robot application described in Section~\ref{sec:application}.  A timelapse of one of our our on-robot experiments is shown in Figure~\ref{fig:app_timelapse} and a video can be found in our supplementary material.  Below we go through the details of the deployment as well as some qualitative observations.

\subsubsection{Human Demonstration of Sorting}
For the deployment of our on-robot application, we asked participants to construct the objects for us from STEM educational construction kits. Some example objects are shown in Figure~\ref{fig:objects}. We then asked the participant to demonstrate a sort by taking a video of themselves moving each object into one of the supplied place locations. Our place locations were a green tray, red tray, white tray, and basket, but any arbitrary locations that can be semantically labeled by a large language model can be used. Multiple novel objects can be placed in the same location. We also asked the human demonstrator to rotate the object twice in front of the camera while placing it to give the manipulated objects detector training pipeline coverage over object viewpoints. An example of a human demonstration can be seen in our supplementary video.  A more comprehensive description of the experience from the participants' perspective can be found in Appendix~\ref{app:demo:part}.

\subsubsection{Qualitative Observations On Robot}
\label{sec:demo:qual}
We executed this application on robot several times with various novel objects, sorting configurations, and object placements, using both VLMs and MOD as the object detector.  While the robot was able to successfully copy human demonstrations involving multiple picks and places when running MOD, we observed two common failure modes of object detection:
\begin{enumerate}[wide=10pt,label=(\alph*)]
    \item Missed Detections: The detector fails to find an object at all.  This resulted in objects not being sorted.
    \item Confusion: The detector confuses one object for another.  This resulted in an incorrect placement of objects in target locations.  With VLMs, these failures occurred both at the semantic category level, e.g., confusing a ``red and white construction toy" with a ``green tray,"  \textit{and} within a category, e.g., confusing a ``red and white construction toy" with a ``yellow and black toy." 
\end{enumerate}
Qualitatively, when MOD was used to detect novel-objects, the rates of missed detections, and inter-object confusion were significantly lowered, consistent with the quantitative evaluation.

%% file: 5-Conclusion.tex
\section{Discussion}
\label{sec:discussion}

\subsection{Limitations}
\label{sec:disc:limitations}
Although the Manipulated Objects Detector outperforms prompting-based VLM approaches, it still requires some degree of human supervision when deployed on a robot (see Appendix~\ref{app:limitations} for a complete list of possible human interventions.).  Missed detections occur when MOD overfits to viewpoints seen in the demonstration and require human teleoperation to reproduce those viewpoints on robot.  Humans are sometimes required to address confusion by removing spurious detections or ``locking in" good ones.  Incorporating image-based mesh reconstruction together with cut-paste–learn~\cite{dwibedi2017cut} style data augmentation and rendering during training could improve both of these issues.

\subsection{Extensions and Future Work}
\label{sec:disc:fw}
In this work we present only a single application of the SODC pipeline, a manipulated objects detector, but in fact this pipeline could be used in multiple ways.  We could visually prompt VLMs, create a ``codebook" of object embeddings, fine-tune a different type of model, etc.  We could also use the SODC pipeline to automatically label manipulated objects in large datasets.

Additionally, in this work, we define “objects of interest” as those directly manipulated by a human. However, the tracking and training components of the SODC and MOD pipeline are agnostic to this choice and can operate on any sparse seeding of object masks or bounding boxes. For example, gaze fixation could serve as an importance signal, enabling the system to include non-interacted but still critical objects.

More broadly, many aspects of our ``Show, Don't Tell" paradigm and our pipeline can be lifted out of the context of object-detection to more general applications. Similar tasks include instance segmentation, human activity parsing, action parameter estimation, etc. 

\section{Conclusion}
\label{sec:conclusion}


We have shown that existing open-set object detectors often struggle with out-of-distribution objects that require complex and unintuitive language descriptions to identify. In contrast, our “Show, Don’t Tell” approach automatically generates labeled datasets by using human–object interactions together with pretrained foundation models.  Detectors trained from this dataset achieve higher accuracy and lower inference latency than large open-set models when recognizing task-relevant objects. We demonstrate the practicality of this approach through a fully integrated on-robot system that learns to sort previously unseen objects from a single human demonstration. Overall, our results suggest that while language remains a powerful modality in vision and robotics, a demonstration can be worth a thousand words. 



%% file: 7-Acknowledgements.tex
\section*{Acknowledgments}
We thank Stefanie Tellex and Dogan Yirmibesoglu for feedback on our paper. Reena Leone, Dawn Wendell, and Carolina Hubbard for feedback on our On-robot application, 
various people with the RAI Institute for their help with data collection for our evaluations. Ekumen for facilitating development of our pipeline, and KeyMakr for evaluation annotations on our dataset. 

%% file: 6-Appendix.tex
\appendix
In our supplementary material, we show the working of our entire pipeline -- demonstration collection, SODC processing, MOD training, and deployment on a real world robot towards our on-robot application. Please see our attached video for a visual depiction of our entire pipeline at work. We present further details of our work in the following appendices:
\begin{itemize}
    \item Appendix~\ref{app:mod}: Algorithmic details, training and parameter details of our SODC \& MOD pipelines
    \item Appendix~\ref{app:plan_skeleton}: Details of how we generate plan skeletons from human videos
    \item Appendix~\ref{app:scenegraph}: Online perception components for the on-robot application
    \item Appendix~\ref{app:impl}: Further implementation details for the on-robot application including skill definitions
    \item Appendix~\ref{app:limitations}: Assumptions and possible operator interventions for the on-robot application
    \item Appendix~\ref{app:demo:part}: Experience of our on-robot application from the perspective of participants
    \item Appendix~\ref{app:objects}: Details of our In-House datasets
    \item Appendix~\ref{app:exp}: Additional evaluations between our proposed SODC / MOD and baseline approaches
    \item Appendix~\ref{app:prompt_tool}: Further details and results on human prompting of VLMs for novel objects
    \item Appendix~\ref{app:gpt_naming_tool}: Further details on automatic prompting of VLMs for novel objects
\end{itemize}

\newcommand{\gp}{\texttt{generate\_parameters}}
\newcommand{\search}{\texttt{Search}}
\newcommand{\grasp}{\texttt{Grasp}}
\newcommand{\place}{\texttt{Place}}


\subsection{Dataset Generation and Training Parameters}
\label{app:mod}
\input{6a-Appendix-Algorithms}
Pseudo-code for the Salient Object Data Creation (SODC) pipeline is shown in Algorithm~\ref{alg:sodc}, and the training pseudo-code for the Manipulated Objects Detector (MOD) is shown in Algorithm \ref{alg:mod_training}.

\subsubsection{Track Clustering Parameters}
We use DBSCAN~\cite{ester1996density} for both spatial and temporal clustering of tracks. For spatial clustering, we define the distance metric as one minus the bounding-box IoU, with a maximum neighborhood distance ($\epsilon$) of 0.4. 
In any given frame, we expect the maximum number of objects that are grasped to be 2 (the demonstrator manipulates at maximum 2 objects at a time, one with each hand). Over the course of the video, we can \textit{conservatively} estimate the maximum number of objects that are manipulated in the video to be the number of frames in the video $\times$ 2 objects per frame, under the (incorrect but conservative) assumption that objects in each frame are independent of one another. 
We therefore set the minimum cluster size that DBSCAN expects to be the number of bounding boxes in the scene divided by twice the number of seed frames used to generate the tracks. We found this value of minimum cluster size to work well in practice. We also tested segmentation IOU for this clustering and found the results to be comparable but the possessing time to be significantly higher. 

For temporal grouping of relabeled tracks, we use one minus the Jaccard score~\cite{jaccard1901distribution} as the distance metric, with a maximum neighborhood distance of 0.4. Similarly to above, we expect there to be at least two tracks from SAMURAI (one forward; one backward) corresponding to a given object, and therefore set the and a minimum cluster size of DBSCAN as 2.  After clustering the tracks, the binary masks within each group are merged on a per-frame basis by retaining pixels that appear in at least at least 50\% of the masks in that group.

\subsubsection{Data Augmentation}
To improve detector robustness on robot, we apply a set of image augmentations using the torchvision transformation library. Images are horizontally flipped with probability 0.5. Color jitter is applied with a magnitude of 0.1 for brightness, contrast, saturation, and hue to introduce mild color variation. Random resized crops are applied with crop sizes between 80\% and 100\% of the original image and aspect ratios between 0.9 and 1.1. A zoom-out transformation is applied with a scale range of 20\% to 100\% of the original image size. Additionally, a mild Gaussian blur is applied using a kernel size of 1 and a sigma of 0.01. To simulate a broader range of camera viewpoints, we also apply random affine and perspective transformations. Perspective transformations use a maximum distortion scale of 0.2. Affine transformations include rotations of up to 15 degrees, translations of up to 10\% of the image size, scaling between 0.9 and 1.1, and shearing with a maximum magnitude of 5 degrees.

These transforms were used for our on-robot application, but were not applied for evaluations (so as to evaluate baselines with more in-distribution images). 

\subsubsection{Training Parameters}
We train MOD using a negative log-likelihood loss and the Adam optimizer with a learning rate of 
$3\times 10^{-4}$ for 10 epochs and a batch size of 10. For faster on-robot demonstrations, we reduce the number of training epochs to 5.

\subsection{Plan Skeletons from Human Videos}
\label{app:plan_skeleton}
As part of the end-to-end application, we generate a full plan skeleton from a human video.  A ``plan skeleton" is a sequence of skills with those parameters that can be generated from only a video or RGBD data of a human performing a task.  For example, from a video of a human picking up a cup and placing it on a table, we can generate the plan [\grasp(``cup"), \place(``table")].  We call skill parameters that can be generated from only the human demonstration ``demonstration time parameters".  The Manipulated Objects Detector is itself a demonstration time parameter.

We further make a distinction between \textit{semantic} and \textit{non-semantic} parameters:
\begin{itemize}
\item \textbf{Semantic parameters}: String type parameters that are a human language phrase.  For example, an object name.
\item \textbf{Non-semantic parameters}: Parameters of any type that are not a natural language phrase.  For example, the manipulated object detector itself or the ID of an object MOD can detect.
\end{itemize}

Each skill type has a unique set of demonstration time parameters.  For example, the parameters of the \search{} skill are the manipulated objects detector and the set of objects for which to search (IDs for objects that can be detected by the manipulated objects detector, names for objects using the semantic detection pipeline).
\subsubsection{Problem Statement}

We define a function $\mathcal{C}\left(I, S\right)$ that takes as input an ordered list $I$ of RGB(D) images and a set of \textit{skills} $S$ and outputs a skill plan $P = \left[(n_k, p_k)\right]$, an ordered list of skill names $n_k$ and instantiated demonstration time parameters $p_k$.  A skill is a tuple $s \coloneq (H, n, d, p, g) \in S$ defined as follows:
\begin{itemize}
\item $H$: A boolean indicating whether the skill can be inferred from a human video.  For example, a robot needs an explicit \search{} skill in its plan skeleton but humans rarely exhibit a behavior we can interpret as ``searching" so $H$ is False for search.
\item $n$: A human-interpretable simple name for the skill
\item $d$: A human-interpretable description of the skill
\item $P \coloneq \{ p \coloneq ({p_\text{name}}, p_\text{description)}\}$: Set of tuple of human interpretable names and descriptions for each of the semantic parameters
\item $g$: A set of types describing non-semantic parameters
\end{itemize}

For example, a semantic definition of {\tt Grasp} is approximately:
\begin{itemize}
\item $H$: True
\item $n$: Pick
\item $d$: Grasp an object once the object has been found
\item $P$: $\{$(Name: \texttt{object\_name}, Description: The object to manipulate)$\}$
\item $g$: [\texttt{int //MOD ID}]
\end{itemize}
Full definitions of all the skills used in the paper can be found in Appendix~\ref{app:impl}

\subsubsection{Method Overview}

\begin{figure*}
\centering
\includegraphics[width=\textwidth]{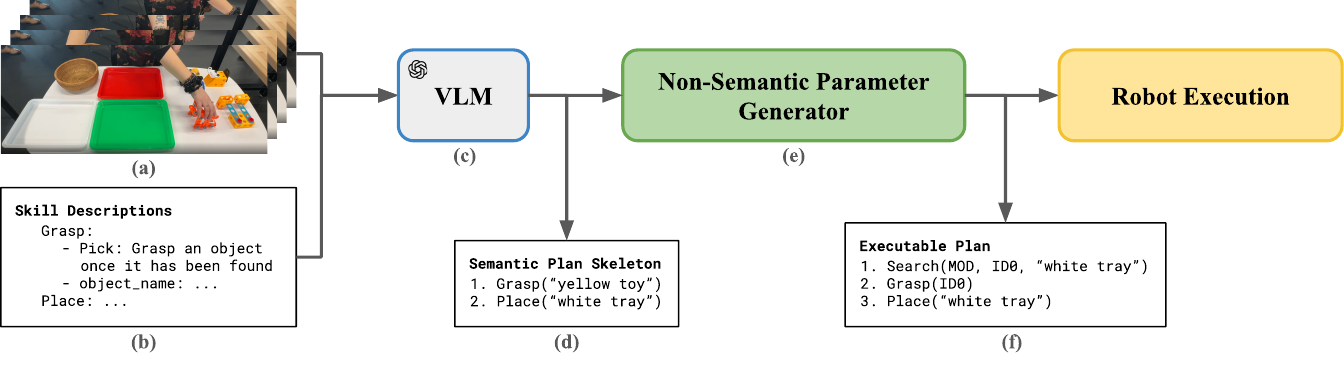}
\caption{The plan skeleton generation pipeline.  Input is the RGB(D) human video (a) and the skills description (b).  The video is sub-sampled into individual image frames and the skills description and images are passed to the LLM (c) with a prompt to generate a plan skeleton.  This semantic plan skeleton (d) is passed to the non-semantic parameter generation (e) which adds any parameters the LLM cannot generate, resulting in the full plan (f).}
\label{fig:copy}
\end{figure*}

If the input is a video (as opposed to a sequence of images), we start by subsampling it into images at about 2hz.  We then generate the plan skeleton in two steps.  We first generate a \textit{semantic plan skeleton} consisting of a subset of skills and their semantic parameters.  We then use that to generate the final plan skeleton.  An overview of the method is shown in Figure~\ref{fig:copy}.

\textit{Semantic Plan Generation} We use an LLM (ChatGPT-4o for the results in this paper) to generate the semantic plan.  We automatically construct a prompt from those skills $s_H = \{s_i \in S \text{ s.t. } H_i \text{ is True}\}$ that might appear in a human video:
\begin{quote}
    You will be provided with a video of a human doing a sequence of $\{n_i ~\forall~ s_i \in s_H \}$.  You should analyze the video to determine the actions taken and the objects involved and return this information using the schema provided.  Here is a description of the actions and the parameters:
    \newline [- $n_i ~: ~d_i~$ \newline 
                \hspace*{0.075in} - [$p_{\text{name}}: p_{\text{description}}]~\forall~ p \in P_i$
                \newline $] ~ \forall s_i \in s_H$
                
     - plan: A list of actions, each of which describes one $\{n_i ~\forall~ s_i \in s_H \}$  along with the objects involved.
\end{quote}
The full prompt used for the paper is shown in Appendix~\ref{app:impl:llm}.

We additionally pass a structured output schema that forces the LLM to return only the skills provided and values for all of their parameters.  We do this by creating a data structure for each skill consisting of string types for each of their arguments.  The structured output requested from the LLM is a list with entries that must be one of the skill types.  The creation of the structured output schema is shown in Algorithm~\ref{alg:structured_output}.
\begin{algorithm}
\caption{Creating the structured output to be passed to the LLM.}
\label{alg:structured_output}
    \begin{algorithmic}[1]
            \State skill\_types $\gets[ \ ]$
            \For{$s\in S_H$}
                \State skill\_types.\texttt{append}($s$.to\_datastructure())
            \EndFor
            \State \texttt{create\_datastructure}(\\
                \hspace{\algorithmicindent}name = ``ActionPlan",\\
                \hspace{\algorithmicindent}plan = \texttt{Field}(\\
                 \hspace{\algorithmicindent}\hspace{\algorithmicindent}type = \texttt{list[Union[skill\_types]]},\\
                \hspace{\algorithmicindent}\hspace{\algorithmicindent}description=``A sequence of actions"\\
                \hspace{\algorithmicindent})\\
                )
    \end{algorithmic}
\end{algorithm}

\textit{Full Plan Generation} Every skill may optionally implement a \gp{}
 function.  This function takes in the demonstration image frames, the current plan skeleton, and the index of the location of the current skill index in the skeleton.  The function can can make arbitrary adjustments to the plan skeleton including adding new skills and changing parameters of any existing skills.  We call this function once per skill instance in a plan.  For example, the \gp{} call for \search{} can start MOD training.  We also use the \grasp{} and \place{} \gp{} function to add \search{} to the plan.  Figure~\ref{fig:copy} shows an example of generating a full plan skeleton from a semantic plan.  Full details of the \gp{} functions we defined for our skills can be found in Appendix~\ref{app:impl}.

\subsection{Online Perception}
\label{app:scenegraph}
\begin{figure*}
\centering
\includegraphics[width=\textwidth]{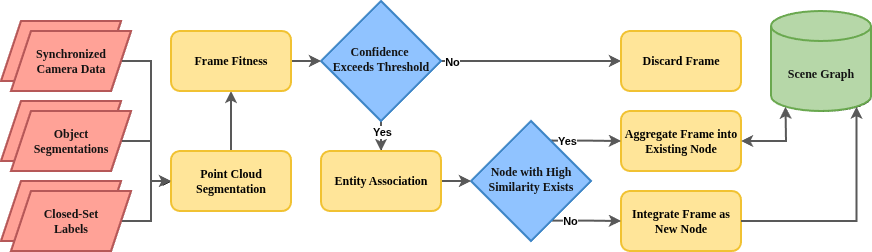}
\caption{The system diagram for the update pipeline of the scene graph. Synchronized camera data input contains RGB images, depth images, intrinsic matrices, and extrinsic matrices. Together with synchronized image frames, the pipeline is also provided pairs of language queries and segmentation masks. These inputs are verified and integrated into the graph as either new nodes or merged with existing nodes.}
\label{fig:scene_graph_update}
\end{figure*}

During on-robot execution, we construct a 3D scene graph that associates and aggregates sensor data and object segmentation masks for a closed set of labels into an object-centric representation of the scene; see Figure \ref{fig:scene_graph_update} for a diagram of the system.
The closed set of labels is the set of objects from the plan skeleton.
These representations are asynchronously updated as the robot moves around and executes a task, minimizing the time required to actively search for task-relevant objects.
Our scene graph encapsulates both geometric and semantic object features.

\subsubsection{Scene Graph Data Structure}
Our scene graph is a set of nodes and edges.
Each node in the graph represents a single object instance in the environment and contains semantic and geometric object information (see Appendix \ref{app:obj_rep} for details).
Edges define relationships or shared features between objects, but for the purpose of this work we only use the graph nodes.

\subsubsection{Object Representation}
\label{app:obj_rep}
Scene graph nodes represent a single instance of an object in the scene.
Some of the attributes of nodes include: object name, 6D pose of the object in the world, segmented point cloud, and segmentation score.

We assume in this work that there is only one object per label in the scene.  Thus, the scene graph should contain a single representation for each task-relevant object in the scene, but, as discussed in Appendix \ref{app:data_association}, this is not guaranteed.
Object information is retrieved from the scene graph using the object name or the MOD ID.
When there are multiple competing representations for a single object instance, the conflict is resolved at query time by ranking the instances by score.
Object score is computed as $S_{obj} = S_{seg} + w * (N_s-1)$ where $S_{seg}$ is the segmentation score of the object, $N_s$ is the number of sightings of the object, and $w$ is a weight on the number of sightings for the score (in practice we set $w=0.02$).

Nodes can have a locking property that, when enabled, guarantees the node will remain the representative instance of the object, regardless of the score of other nodes of the same label.  As discussed in Appendix~\ref{app:limitations}, the operator can toggle locking for a node.

\subsubsection{Object Detection and Segmentation}
In addition to synchronized camera data, our scene graph also requires object segmentations to integrate new object information.
Our pipeline uses GroundingDINO~\cite{liu2023groundingdino} or our manipulated objects detector for detection and SAM2~\cite{ravi2024sam} for segmentation.

\subsubsection{Frame Fitness}
For new input data to be integrated into the scene graph, it must pass a confidence-based frame fitness check that considers the object's segmentation score and a per-pixel confidence score.

The per-pixel confidence score is computed by taking the segmentation score and inversely scaling it with the square root of the number of points in the object's segmented point cloud: $S_{pix}=\frac{S_{seg}}{\sqrt{N_{p}}}\alpha$, where $S_{seg}$ is a segmentation score associated with the frame, $N_{p}$ is the number of points in the segmented point cloud, and $\alpha$ is a scalar. 
This score weighs point clouds with more points as higher risk and requires their confidence to be higher to be accepted into the scene graph.

The frame is processed if the segmentation score and the per-pixel confidence score are sufficient, otherwise, it is discarded.
In practice, we use a value of $\alpha=1000$, $0.3$ as the segmentation score threshold, and $10.0$ as the per-pixel confidence threshold.  These values were tuned empirically.

\subsubsection{Acceptance Groups}
\label{app:group_acceptance}
In addition to enforcing frame fitness, we can also impose \textit{group acceptance} on subsets of detection labels such that all labels associated with the group must be detected within a specified time window to be integrated into the scene graph.

Our manipulated objects detector performs best when all relevant objects are visible simultaneously.
On robot, we enforce this condition on the scene graph by creating an acceptance group containing all MOD-relevant objects with a time window of zero for each MOD instance.
This has been observed to reduce confusion among MOD objects, but relies on the MOD instance being capable of detecting all objects consistently.  As discussed in Appendix~\ref{app:limitations}, the operator can toggle this group acceptance on and off.

\subsubsection{Entity Association and Frame Integration}
\label{app:data_association}
Before a frame is integrated into the scene graph, an entity association algorithm is run to determine whether the data should be inserted as a new node or be merged into an existing, reference node. 

For each existing reference node that shares a name with the candidate node, we compute the fraction of nearest neighbors between the candidate and reference point clouds.
The candidate is merged into the node with the highest overlap percentage if that percentage is above a threshold.
Otherwise, the node is integrated into the graph separately as a new instance of the label type.

It's worth noting that it is possible for multiple nodes representing the same object label to exist if this algorithm does not find sufficient similarity between them; this motivates our object scoring at query time, discussed in Appendix \ref{app:obj_rep}.

\subsection{Implementation Details}
\label{app:impl}

For the application described, we implemented three skills: \search, \grasp, and \place.  We describe each in detail including their implementation on the Boston Dynamics Spot Robot.

\subsubsection{Search}

\mbox{}

\textit{Semantic Description}: We do not consider \search{} a skill that can be observed from human videos.  Therefore, it does not require a semantic description.

\textit{Generate Parameters}: The \gp{} function for search starts MOD training if requested and stores the resulting checkpoint data.  In our application, we started MOD training in parallel to plan generation so this functionality was not used.

\textit{Execution}: At the start of execution, \search{} matches grasped objects to their MOD IDs.  This is currently done with the simple algorithm of assuming that the order in which objects are grasped matches their IDs (so the first grasped object is MOD ID 0 and so on).  This can fail if MOD finds sub-parts for an object so we also allow human intervention to map grasped objects to MOD IDs.  We additionally allow a human to intervene to tell \search{} to skip searching for some objects or to replace object names with other object names.  The object labels are sent to the scene graph and that becomes the closed set of objects over which the scene graph operates.

\search{} executes on robot in two phases:
\begin{enumerate}
    \item Autonomous search scans the environment autonomously.  It draws a padded bounding box around currently detected objects and sweeps the floor using its body and arm to ensure that the gripper camera sees the whole bounding box.  The bounding box is enlarged each time new objects are found.  The box is seeded with objects found during an initial sweep of the arm in front of the robot.  To prevent placing on the edge of an object, any objects on which the robot will place are only considered ``found" if their x and y dimensions both exceed 0.2m.
    \item Once the robot believes it has found all objects, or if nothing is found during the initial sweep, or if the robot has done a maximum number of sweeps (2 in practice), it will ask the human if they want to teleoperate to find more objects.  The human can then teleoperate the robot at will until all objects are found.
\end{enumerate}

Throughout \search{}, humans can intervene to change the objects for which the robot is looking, toggle whether objects are ``locked" (Appendix~\ref{app:obj_rep}), and toggle whether group acceptance is used (Appendix~\ref{app:group_acceptance}).  At the end of the execution, \search{} offers the human a chance to adjust the plan for what was found.  For a full list of potential human interventions see Appendix~\ref{app:limitations}.

\subsubsection{Grasp}

\mbox{}

\textit{Semantic Description}: The semantic descriptions for \grasp{} is:
\begin{itemize}
    \item Name: Pick
    \item Description: ``Grasp an object once the object has been found"
    \item Parameters: \begin{itemize}
        \item \texttt{object\_name}: ``The object to manipulate"
        \item \texttt{future\_manipulation}: ``The next action we will do with the object after picking it up"
    \end{itemize}
\end{itemize}
The ``future\_manipulation" parameter is likely unneeded but was left in the implementation from previous experiments.

\textit{Generate Parameters}: The \gp{} function for \grasp{} adds the \search{} skill to the beginning of the plan if it isn't already there and adds the object name to the set of objects for which to search.

\textit{Execution}: At the start of the grasp, we position the robot to gaze at the object from a pose above and slightly offset using the pose of the object from the scene graph.  We implemented two methods of grasping on the robot:
\begin{enumerate}
    \item Spot API: We pass a ray to the object to the Spot API grasping call and allow Spot to execute the grasp if it can.
    \item GraspGen: If the Spot API is unable to execute a grasp, we use the aggregated point cloud of the object from the scene graph and plan a grasp using GraspGen~\cite{graspgen}.  We execute the planned grasp via the Spot API end effector Cartesian setpoints.
\end{enumerate}
If after grasping the robot senses that its gripper is shut and a human confirms the grasp was missed, it retracts up and attempts to re-acquire and redo the grasp.

\subsubsection{Place}

\mbox{}

\textit{Semantic Description}  The semantic description for \place{} is:
\begin{itemize}
    \item Name: Place
    \item Description: ``Place an object onto another object.  Cannot be used for putting an object carefully inside another or into a hole."
    \item Parameters: \begin{itemize}
        \item \texttt{place\_name}: ``Where to put the object currently being held"
        \item \texttt{careful}: ``Whether to carefully place the object down or drop it"
    \end{itemize}
\end{itemize}
The LLM never chose to set \texttt{careful} False.

\textit{Execution} We choose the center of the bounding box of the object on which to place as the place location for the bottom of the grasped object.  We always place with the gripper in a top down pose and calculate the height of the gripper at place based on the height of the object being held (or at a constant offset if \texttt{careful} was False but this never happened in practice).  We use the Spot API end effector Cartesian setpoint control to move the robot to this pose and open the gripper.

\subsubsection{Full LLM Prompt}
\label{app:impl:llm}
The semantic descriptions of \grasp{} and \place{} are used to automatically generate the LLM prompt as discussed in Appendix~\ref{app:plan_skeleton}.  For the application described in the paper, the generated prompt was:

\begin{framed}

\begin{quote}
    You will be provided with a video of a human doing a sequence of places and picks.  You should analyze the video to determine the actions taken and the objects involved and return this information using the schema provided.  Here is a description of the actions and the parameters:\newline
	- Place: Place an object onto another object.  Cannot be used for putting an object carefully inside another or into a hole.\newline
	\hspace*{0.075in} - place\_name: Where to put the object currently being held \newline
	\hspace*{0.075in} - careful: Whether to carefully place the object down or drop it \newline
	- Pick: Grasp an object once the object has been found\newline
	\hspace*{0.075in} - object\_name: The object to manipulate\newline
	\hspace*{0.075in} - future\_manipulation: The next action we will do with the object after picking it up\newline
	- plan: A list of actions, each of which describes one place or pick along with the objects involved.
\end{quote}
\end{framed}

\subsection{Assumptions and List of Potential Human Interventions}
\label{app:limitations}
We made the following assumptions to simplify copying a human video:
\begin{itemize}
    \item Humans touch each object only once (this was specified in instructions to demonstrators).
    \item Only one instance of each object will be in the robot's scene during execution.
\end{itemize}

We allow several possible operator interventions.  It is possible for the robot to succeed with none of these interventions but having them makes successful execution more likely:
\begin{itemize}
    \item Correct mapping of grasped objects to MOD IDs.
    \item Teleoperate robot until all objects are detected during \texttt{Search} skill.
    \item Ask the robot to forget all detections of an object
    \item Tell robot to ``lock" a detection so that the label cannot move to another object.
    \item Toggle whether to force the robot to see all MOD objects to detect them (``group acceptance")
    \item Replace the labels or MOD IDs in the plan with new names
    \item Skip a step in the plan
    \item Verify whether or not a grasp was successful if the robot's gripper is fully closed.
    \item Teleoperate the robot to find the object to grasp again after a missed grasp if the robot was unable to find it itself.
\end{itemize}

\subsection{On Robot Application Participants}
\begin{figure}
    \centering
    \includegraphics[width=\linewidth]{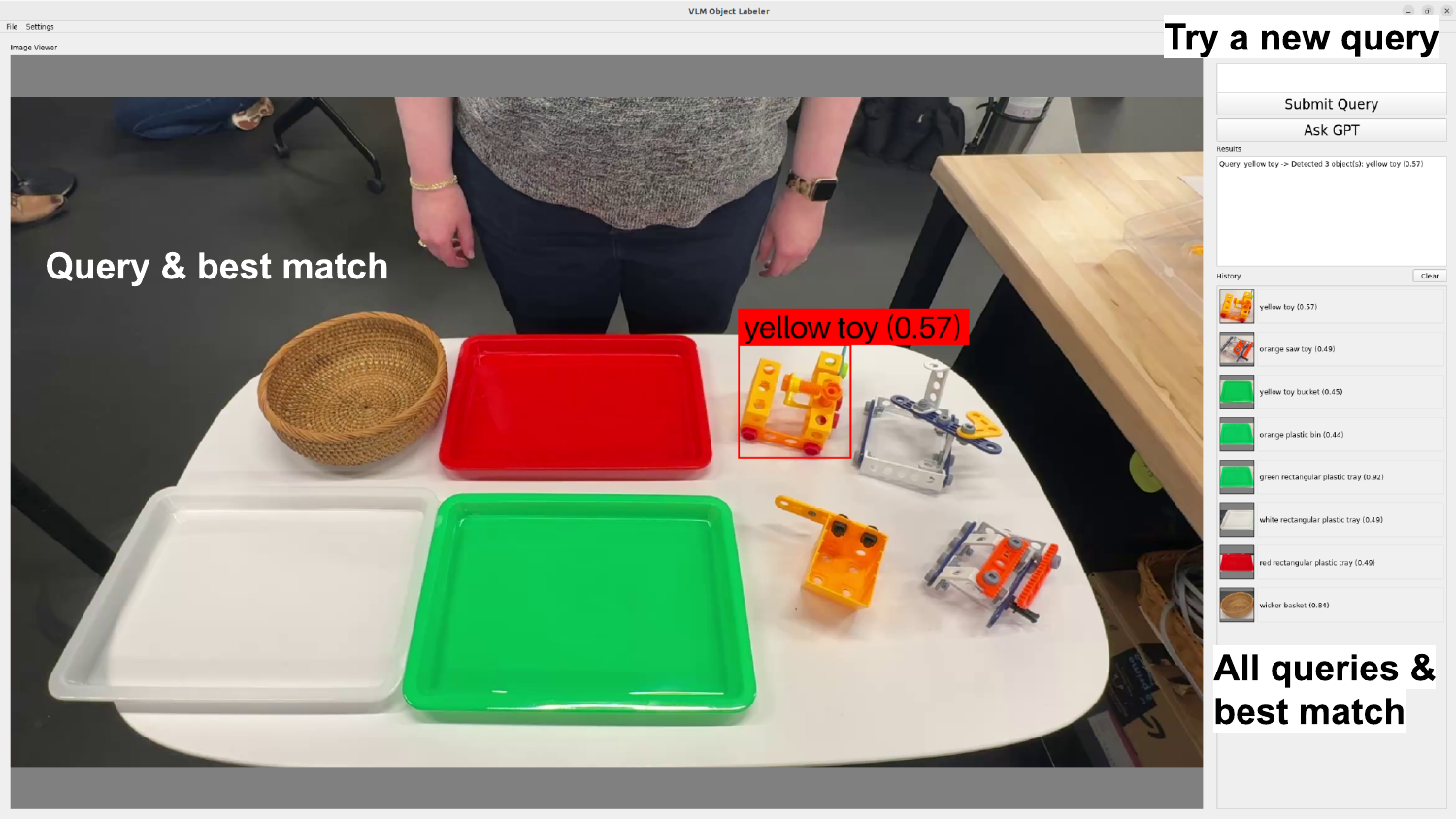}
    \caption{The ``name game" where participants can try different prompts for the objects they have built.}
    \label{fig:namegame}
\end{figure}
\label{app:demo:part}
From the point of view of participants in the on-robot application described in Section~\ref{sec:exp:application}, we begin the experience with a ``cold open".  Each participant is given a different construction kit and instructed:
\begin{quote}
Please spend 5 minutes building a small object that Spot can grasp easily.  You should test that Spot can grasp it by placing it on the table and using the Spot gripper provided to pick it up with a top down grasp.  Please make the object rigid and difficult to break as it will eventually be manipulated by spot. I will let you know when 5 minutes are coming up so you can put some finishing touches on the creation. You can use the tools to tighten up bolts if you feel like something is loose.    
\end{quote}

One participant is then chosen to show the robot how to sort the objects sort into the provided place locations.  We film the demonstrated sort with a phone camera.  The instructions given to the participant are:
\begin{quote}
    Please pickup objects, show them to the camera by rotating them two times, and place them in a tray, including your created object. You do not need to sort all objects. Please move each object only once, and use only one hand at a time. You can use either hand but only one at a time. We will take a 15 second video of this. I will count down 3.. 2.. 1.. as the time is finishing up.
\end{quote}

Once the filming of the participant sorting objects is finished, the robot has all the data it needs to start the pipeline described in Section~\ref{sec:exp:application}.  While the robot processes, we explain the system to the participants and invite them to try the ``Name Game".  This is an interactive tool in which participants craft prompts describing their novel objects and we show the best detection of a VLM as shown in Figure~\ref{fig:namegame}.  This experience motivates our creation of the manipulated objects detector by showing participants how difficult it can be to prompt VLMs to recognize objects.  We additionally show participants the automatically generated training data of their own novel objects to emphasize how quickly the training data can be generated and enforce the overall understanding of the pipeline.  During this time, the robot also conducts a search for the manipulated objects and the place locations.

Finally, participants watch as the robot copies the sort they demonstrated.

\subsection{In-House Datasets}
\label{app:objects}
In order to develop significant evaluations of our pipeline, it was necessary to collect datasets with human-hand interactions with abstract objects.  We presented these two datasets in Section~\ref{sec:exp:setup} and here give more details.  Videos with samples from both of the datasets can be found in the supplemental material submitted.

\subsubsection{Dataset Properties}
We show the properties of each dataset in Table~\ref{tab:in_house_datasets}.  The bold properties show the parameters each dataset was focused on showcasing.  The property definitions are:
\begin{itemize}
    \item \textit{\# Objects: }The number of distinct objects over all videos.
    \item \textit{Common Objects: }Both datasets have objects made from STEM construction kits.  In one dataset we also included more commonly available but unusual objects (toys, tools, etc).
    \item \textit{\# Object Creators: }Number of participants involved in the creation of the novel objects. This ensured wider breadth of creative features in constructed objects.
    \item \textit{\# Object Sets: }Number of different combinations of object sets across videos. All videos included multiple objects.
    \item \textit{Conditions: }Number of additional objects in the field-of-view of the camera, from:
        \begin{enumerate}[label=\alph*)]
            \item \textit{None}: Only objects of interest in frame.
            \item \textit{Trays}: Objects with baskets and trays.
            \item \textit{Distractors}: Objects from other object sets included as distrators.
        \end{enumerate} 
    \item \textit{Interactions: }Type of interactions with the objects of interest from:
        \begin{enumerate}[label=\alph*)]
            \item \textit{Sorting}: Each object is grasped and moved to a tray.
            \item \textit{Freeform}: Objects are interacted with randomly.
            \item \textit{Partial}: Not all objects are grasped and moved.
            \item \textit{Simultaneous}: Both hands are used to grab and move objects simultaneously.
            \item \textit{Sequential}: Both hands are used to move objects, but only one hand at a time.        
        \end{enumerate} 
    \item \textit{\# Backgrounds: } Number of different backgrounds used in the videos.
    \item \textit{\# Demonstrators: }Number of different participants performing the interactions.
    \item \textit{\# Robot POV Videos: }How many of the videos in the dataset are taken from the robot in-hand camera.
    \item \textit{\# VLM Prompt Annotations: }Number of human-farmed annotations (2 per video, max 5 attempts per object) as described in Appendix~\ref{app:prompt_tool}.
\end{itemize}

\begin{table}[ht]
    \centering
    \caption{Properties of In-House Datasets}
    \label{tab:in_house_datasets}
    \resizebox{\columnwidth}{!}{
    \begin{tabular}{@{}lcc@{}}
        \toprule
        \textbf{Property} & \textbf{Dataset 1} & \textbf{Dataset 2} \\
        \midrule
        \# Objects & 12 & 17\\
        Common Objects & Yes & No\\
        \# Object Sets & 3 & \textbf{12}\\
        \# Object Creators & 6 & \textbf{12}\\
        Conditions & a,b,c & a,b\\
        Interactions & a,b,c,d,e & a,b\\
        \# Backgrounds & \textbf{4} & 1\\
        \# Demonstrators & 1 & \textbf{17}\\
        \# Robot POV Videos & \textbf{18} & 0\\
        \# VLM Prompt Annotations & \textbf{108} & 0\\
        Total Videos & 72 & 61\\
        \bottomrule
    \end{tabular}
    }
\end{table}

\subsubsection{Dataset Differences} 
Both in-house datasets are complimentary given the selected parameters, providing good coverage for a variety of objects and conditions. 

\textit{Dataset 1 - In-House Human and Robot: }Contains robot POV videos. and baseline non-constructed (common) objects. Annotated with human-derived VLM prompts. Additional backgrounds and environment conditions.

\textit{Dataset 2 - In-House Human} Contains novel/constructed objects only. Multiple participants performing the demonstrations. Larger number of distinct objects and objects sets.

\subsection{Additional Evaluations}
\label{app:exp}
\input{4-2-table2}
In addition to the evaluations presented in our main paper, we also carry out additional evaluations described below. 

\subsubsection{Motivation from Human Prompt Farming}
The human prompt farming study (Appendix~\ref{app:prompt_tool}) provides strong empirical motivation for our ``Show, Don't Tell'' approach. When humans attempted to craft prompts for novel construction-kit objects, the first-attempt success rate was only 20.6\%, and even after five attempts, 43\% of objects could not be detected by open-vocabulary detectors. This stands in stark contrast to common objects like ``mug'' or ``drill,'' which achieved near-perfect detection with single-word prompts. These results demonstrate that prompt engineering for novel objects is not merely tedious---it is fundamentally unreliable, providing further justification for bypassing language-based detection entirely in favor of our self-supervised MOD approach.



\subsubsection{Supervised MOD Baseline}
In order to test out the efficacy of our proposed Salient Object Dataset Creation (SODC) pipeline, we evaluate an additional baseline that trains an object detector identical in architecture to our Manipulated Objects Detector (MOD), on the true ground truth data available on each of the datasets used.

We do so by carrying out evaluations in a ``leave-one-out'' fashion, where we train a single object detector (a supervised MOD) on a single video from the test set of a given dataset using the ground truth annotated images provided rather than the ones generated by SODC.  We evaluate the metric performance of this supervised MOD on \textit{each} other video present in the test set of that dataset. We then carry this out for each of the datasets presented.  We refer to this baseline as the ``Supervised MOD" baseline, and present evaluations for it in Table~\ref{table:supervisedmod}. 

In principle, a supervised MOD trained on the ground truth dataset should serve as an \textit{upper bound} in performance to our own SODC + MOD combination. 
However, we note that in practice, ground truth data may be sparse in time, i.e., not all frames of a given video may be annotated. In contrast, our SODC constructs datasets that are temporally dense, so it is possible for us to observe SODC + MOD performance on par with, or higher than that of the supervised MOD baseline. 

For our In House dataset, we observe in Table~\ref{table:supervisedmod} that the supervised MOD baseline only marginally outperforms our SODC + MOD approach, despite being trained on the ground truth annotations. 
In the Meccano dataset, we observe that our SODC + MOD approach actually outperforms the supervised MOD baseline. Our hypothesis for this is that annotations for the Meccano dataset are sparse in time for the objects being detected. As a result, the supervised MOD is trained on a smaller, less representative dataset than our own SODC dataset, which automatically creates temporally dense data (by virtue of the tracking component used). 
Collectively, the success of our SODC + MOD approach to either be as performant as, or indeed, outperform the supervised MOD with ground truth annotations is strong empirical evidence to the quality of datasets generated by our SODC. 


\subsection{Human Prompt Farming}
\label{app:prompt_tool}

To quantify the effort required for humans to craft effective prompts for open-vocabulary detectors, we developed an interactive annotation tool and conducted a study with 16 participants.

\subsubsection{Annotation Tool}
We built a PyQt5-based GUI application that allows users to:
\begin{enumerate}
    \item Load video files (of human demonstrations assigned to participant) and navigate frames.
    \item Draw a tight bounding box around a target object in the video frame.
    \item Enter a natural language prompt intended to make the detector localize that object.
    \item Receive immediate feedback on whether the prompt succeeded.
\end{enumerate}
Each prompt is evaluated simultaneously by two open-vocabulary detectors: GroundingDINO~\cite{liu2023groundingdino} and Detic~\cite{zhou2022detecting}. For each detector, the tool computes the IoU between the human-drawn bounding box and the detector's highest-confidence predicted bounding box for the given prompt. An attempt is considered \textit{successful} if at least one detector achieves $\mathrm{IoU}~\geq 0.5$. Participants are given a maximum of five attempts per object before moving on.

The tool records for each attempt: the prompt text, per-detector detection results (detected/not, confidence, predicted bounding box, IoU), and which detector(s) agreed.

\subsubsection{Procedure}
Participants are given an ordered list of videos from In-House Dataset~1 to annotate. Each video contained 4--6 task-relevant objects. Participants were instructed to:
\begin{itemize}
    \item Draw tight bounding boxes around each object (as IoU determines success).
    \item Approach each object ``naively,'' using only descriptions of the object itself---positional prompts (\textit{e.g.}, ``the object near the human's hand'') were prohibited.
    \item Annotate videos in the assigned order, to enable analysis of learning effects.
    \item Move on after either achieving a successful detection or exhausting all five attempts.
\end{itemize}
In total, participants annotated 507 object instances across 12 distinct object types. Of these 12 objects, six were novel objects assembled from Meccano construction kits (the objects shown in Figure~\ref{fig:objects}
), one was a 3D-printed part, and five were common household or tool objects (\textit{e.g.}, a mug, a drill, a toy cat).

\subsubsection{Results}

\paragraph{Overall statistics}
Across all 507 objects instances and 1,136 prompt attempts, 78.3\% of objects were successfully detected within five tries, with a first-attempt success rate of 55.2\%. However, performance differed dramatically between common and novel objects.

\paragraph{Common vs.\ novel objects}
As shown in Table~\ref{tab:prompt_farming}, common real-world objects (Cat, Drill, Mug, Screwdriver, YellowTruck) had a near-perfect 99.5\% success rate, with 92.4\% of attempts succeeding on the first try. In contrast, the six novel construction-kit objects had only a 57.0\% success rate, with a 43.0\% failure rate and an average of 3.25 attempts per object (including objects that exhausted all five attempts without success). This stark difference---from near-trivial prompting for known object categories to highly error-prone prompting for novel objects---underscores the core motivation of our ``Show, Don't Tell'' approach.

\begin{table}[htbp]
    \centering
    \caption{Per-object prompt farming results. Success rate is the fraction of object instances for which at least one prompt succeeded within 5 attempts. Mean attempts is the average number of attempts used per object (including objects that exhausted all 5 attempts without success). Objects are grouped into common (top) and novel construction-kit (bottom) categories.}
    \label{tab:prompt_farming}
    \resizebox{\columnwidth}{!}{
    \begin{tabular}{@{}lccccc@{}}
        \toprule
        \textbf{Object ID} & \textbf{Trials} & \textbf{Success (\%)} & \textbf{Fail (\%)} & \textbf{Mean Att.} & \textbf{1st Att.} (\%) \\
        \midrule
        Cat & 32 & 100.0 & 0.0 & 1.00 & 100.0 \\
        Drill & 47 & 100.0 & 0.0 & 1.04 & 95.7 \\
        Mug & 48 & 100.0 & 0.0 & 1.04 & 95.8 \\
        Screwdriver & 44 & 97.7 & 2.3 & 1.19 & 83.7 \\
        YellowTruck & 40 & 100.0 & 0.0 & 1.10 & 90.0 \\
        \midrule
        \textit{Common avg.} & \textit{211} & \textit{99.5} & \textit{0.5} & \textit{1.09} & \textit{92.4} \\
        \midrule
        BlackYellowBox & 42 & 61.9 & 38.1 & 3.36 & 19.2 \\
        GooglyEyedToy & 49 & 55.1 & 44.9 & 3.27 & 22.2 \\
        OrangeBracket & 42 & 47.6 & 52.4 & 3.45 & 10.0 \\
        Sawzall & 43 & 60.5 & 39.5 & 3.19 & 19.2 \\
        SnowShoe & 36 & 55.6 & 44.4 & 3.50 & 25.0 \\
        YellowDelta & 37 & 62.2 & 37.8 & 2.73 & 30.4 \\
        \midrule
        \textit{Novel avg.} & \textit{249} & \textit{57.0} & \textit{43.0} & \textit{3.25} & \textit{20.6} \\
        \midrule
        PrintedPart & 47 & 95.7 & 4.3 & 1.98 & 53.3 \\
        \bottomrule
    \end{tabular}
    }
\end{table}

\paragraph{Detector agreement}
The two detectors were complementary: across all attempts, GroundingDINO succeeded on 11.6\% of attempts, Detic alone on 11.5\%, and both agreed on 11.8\%, while neither succeeded on 65.1\%. The union of the two detectors yielded a per-attempt success rate of 34.9\%, substantially higher than either detector individually ($\sim$23\% each). This indicates that using multiple open-vocabulary detectors can improve prompt-based detection, but even the union leaves a high failure rate on novel objects.

\paragraph{Prompt characteristics}
Participants used significantly longer prompts for novel objects (4.3 words on average) than common objects (2.7 words), reflecting their attempts to describe objects that lack simple category names. However, within novel objects, longer prompts did not improve success: failed prompts averaged 4.4 words while successful prompts averaged only 3.6 words. This suggests that ``trying harder'' with more detailed descriptions does not help---and may even hurt---detection performance. The successful prompts for novel objects tended to be creative metaphorical mappings to known visual concepts (\textit{e.g.}, ``sled,'' ``claw,'' ``shoe'') rather than literal descriptions (\textit{e.g.}, ``multi-part construction toy with gray, white, and yellow parts''). This makes prompt engineering for novel objects more of an art: success depends on guessing which visual metaphor might resonate with the detector's training data, a fundamentally unreliable process.

\paragraph{Participant variability}
Per-participant success rates ranged from 67.9\% to 90.6\% (std.\ dev.\ 6.6\%), suggesting that prompt crafting ability varies across individuals but remains challenging for all participants on novel objects.

\paragraph{Learning effects}
We observed a mild improvement of $\sim$2\% in success rate between each participant's first and second halves of annotations (9 of 16 participants improved, 4 declined, 3 were unchanged). The small effect size suggests that participants do not substantially improve at prompt crafting with practice over a single session, reinforcing that the difficulty is intrinsic to the task rather than a matter of user experience.

\paragraph{Implications}
These results reveal a fundamental limitation of prompt-based detection for novel objects. When humans---who can see the object, understand the task context, and iteratively refine their descriptions---fail to produce a working prompt 43\% of the time even after five attempts, it is unreasonable to expect automated prompt generation (\textit{e.g.}, via GPT) to reliably succeed. The problem is not that humans are bad at prompting; it is that the objects themselves lack the semantic anchors that VLMs require. A custom Meccano assembly has no established name, no canonical description, and no visual prototype in the detector's training data. Our ``Show, Don't Tell'' approach sidesteps this problem entirely: rather than struggling to describe that which is difficult to describe, we simply show the detector what to look for by training on automatically-generated examples from the demonstration video.

\subsection{Automatically Generating Prompts}
\label{app:gpt_naming_tool}
\begin{figure}[htbp]
    \centering

    \includegraphics[width=0.09\textwidth]{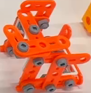}
    \includegraphics[width=0.09\textwidth]{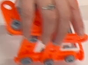}
    \includegraphics[width=0.09\textwidth]{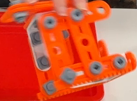}
    \includegraphics[width=0.09\textwidth]{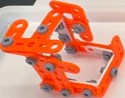}
    \includegraphics[width=0.09\textwidth]{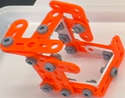}
    \includegraphics[width=0.09\textwidth]{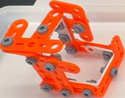}
    \includegraphics[width=0.09\textwidth]{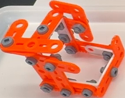}
    \includegraphics[width=0.09\textwidth]{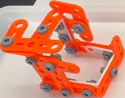}
    \includegraphics[width=0.09\textwidth]{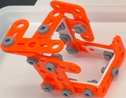}
    \includegraphics[width=0.09\textwidth]{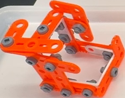}
    \textbf{Object 0}
    
    \vspace{0.2cm}

    \includegraphics[width=0.09\textwidth]{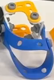}
    \includegraphics[width=0.09\textwidth]{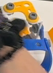}
    \includegraphics[width=0.09\textwidth]{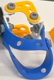}
    \includegraphics[width=0.09\textwidth]{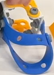}
    \includegraphics[width=0.09\textwidth]{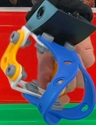}
    \includegraphics[width=0.09\textwidth]{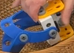}
    \includegraphics[width=0.09\textwidth]{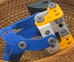}
    \includegraphics[width=0.09\textwidth]{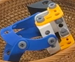}
    \includegraphics[width=0.09\textwidth]{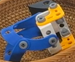}
    \includegraphics[width=0.09\textwidth]{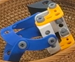}
    \textbf{Object 1}
    
    \vspace{0.2cm}
    
    \includegraphics[width=0.09\textwidth]{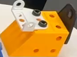}
    \includegraphics[width=0.09\textwidth]{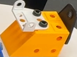}
    \includegraphics[width=0.09\textwidth]{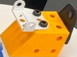}
    \includegraphics[width=0.09\textwidth]{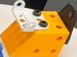}
    \includegraphics[width=0.09\textwidth]{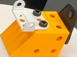}
    \includegraphics[width=0.09\textwidth]{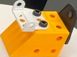}
    \includegraphics[width=0.09\textwidth]{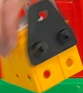}
    \includegraphics[width=0.09\textwidth]{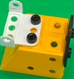}
    \includegraphics[width=0.09\textwidth]{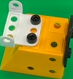}
    \includegraphics[width=0.09\textwidth]{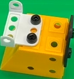}
    \textbf{Object 2}
    
    \vspace{0.2cm}
    
    \includegraphics[width=0.09\textwidth]{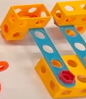}
    \includegraphics[width=0.09\textwidth]{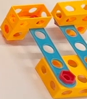}
    \includegraphics[width=0.09\textwidth]{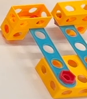}
    \includegraphics[width=0.09\textwidth]{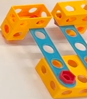}
    \includegraphics[width=0.09\textwidth]{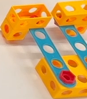}
    \includegraphics[width=0.09\textwidth]{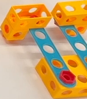}
    \includegraphics[width=0.09\textwidth]{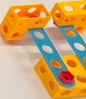}
    \includegraphics[width=0.09\textwidth]{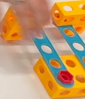}
    \includegraphics[width=0.09\textwidth]{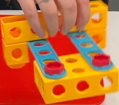}
    \includegraphics[width=0.09\textwidth]{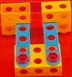}
    \textbf{Object 3}
    
    \caption{Label images cropped by bounding box}
    \label{fig:label_images}
\end{figure}

In order to create the ``GPT prompts" from Section~\ref{sec:exp:promptgen} for our datasets, we wrote a tool that uses ChatGPT-4o to name objects. The tool loads the training data set (generated by SODC) and extracts images for each ID at regular intervals over the duration of the training set, up to a maximum of 10 as shown in Figure~\ref{fig:label_images}. The tool uses the following prompt:
\begin{framed}
\begin{quote}
``Create object query prompts for the objects shown in these images. The prompts will be used by GroundingDino. Do not include locations in the image in the prompts.\newline
\newline
There are 4 different objects. Each object has multiple example images grouped together. For each object, create a single descriptive object query prompt."\newline
\newline
The objects are: 0, 1, 2, 3
\end{quote}
\end{framed}

Example output:

\begin{framed}
\begin{quote}
0: ``bright orange construction toy assembly made of multiple slotted plates joined with many gray circular connectors, forming an angular open-frame structure with white brackets"\newline
1: ``blue curved plastic strap loop with circular holes connected to a yellow slotted bracket and white block, assembled with gray round connectors and a small black piece"\newline
2: ``yellow wedge-shaped block with round holes attached to a white perforated L-bracket with black round knobs and a black triangular plate"\newline
3: ``construction toy arrangement of four yellow rectangular hole blocks connected by two parallel blue slotted strips with two pink circular caps"\newline
\end{quote}
\end{framed}




%% file: 6a-Appendix-Algorithms.tex
\begin{algorithm*}[!]
\caption{Salient Objects Dataset Creation (SODC) Pipeline}
\label{alg:sodc}
\begin{algorithmic}[1]
\Require Video $\mathcal{V}$ of human-object interaction
\Ensure Dataset $\mathcal{D}$ of images with labeled bounding boxes

        

    \item[] \Statex \textbf{Step 1: Detect Grasped Entities}
    \State Initialize $\mathcal{M}_{\text{grasp}} = \{\}$ 
    \Comment{$\mathcal{M}_{\text{grasp}}$ is an empty set of grasp masks.}    
    \State $\mathcal{M}_{\text{grasp}} \leftarrow \text{HOIST-Former}(\mathcal{V})$        
        \Comment{{$\mathcal{M}_{\text{grasp}}$ is a set of segmentation  masks  for frames where an object is grasped.}}

    \item[] \Statex \textbf{Step 2: Track Grasp Masks across Video}
    \State Initialize $\mathcal{T}_{\text{mask}} = \{\}$
    \Comment{$\mathcal{T}_{\text{mask}}$ is an empty set of tracks.}
    
    \For{Grasp Mask $m_{\text{grasp}} \in \mathcal{M}_{\text{grasp}}$}
        
        \State $\mathcal{T}^m_{\text{forward}}, \mathcal{T}^m_{\text{backward}}  \leftarrow \text{SAMURAI}(m_{\text{grasp}}, \mathcal{V})$
        \multicomment{$\mathcal{T}^m_{\text{forward}}, \mathcal{T}^m_{\text{backward}}$ are forward and backward \\ SAMURAI tracks of grasp mask $m$ across $\mathcal{V}$.}
         \item[]
        \State $\mathcal{T}_{\text{mask}} = \mathcal{T}_{\text{mask}} \cup \  \{ \mathcal{T}^m_{\text{forward}}, \mathcal{T}^m_{\text{backward}} \} $
        \Comment{Aggregate tracks across grasp masks.}   
    \EndFor

    \item[]
    \State $\mathcal{T}_{\text{bbox}} \leftarrow \text{MaskToBBox}(\mathcal{T}_{\text{mask}})$
    \Comment{Convert mask tracks $\mathcal{T}_{\text{mask}}$ to bounding box tracks $\mathcal{T}_{\text{bbox}}$.}
    \item[] \Statex \textbf{Step 3a: Spatial Bounding Box Clustering}
    \State Initialize $\mathcal{C}_{\text{spatial}}  = \{\}$    
    \Comment{$\mathcal{C}_{\text{spatial}} $ is an empty set of per-frame spatial clusters.}
    \For{Timestep $t \in \mathcal{V}$}
        \State $B_t \leftarrow$ Bounding Boxes from $\mathcal{T}_{\text{bbox}}$ at time $t$
        \Comment{Retrieve set of all tracked bounding boxes at time $t$.}
        \State $\text{IoU}_t \leftarrow \text{IoU}(b_i, b_j) \ \forall \ b_i, b_j \in B_t$ 
        \Comment{Compute IoU between each pair of boxes in $B_t$.}
        \State $ \{ F_t C_b, b \}_{b \in {B_t}} \leftarrow$ DBSCAN($B_t$, $\text{IoU}_t$) 
        \multicomment{Cluster bounding boxes $b$ in $B_t$ at time $t$ based on  IoU\\ distance with cluster labels $F_t C_b$ for box $b$ for frame $F_t$.}
        \For{Bounding Box $b \in B_t$}
            \State $ \mathcal{C}_{\text{spatial}} = \mathcal{C}_{\text{spatial}} \cup \{ F_t C_b, b \}$
            \Comment{Aggregate spatial cluster labels across frames in the video.}
        \EndFor
        
    \EndFor
    

    \item[] \Statex \textbf{Step 3b: Build Cluster Tracks}
    \State Initialize $\mathcal{C}_{\text{tracks}}  = \{\}$
    \Comment{$\mathcal{C}_{\text{tracks}}$ is an empty graph of tracks to be clustered.}
    \For{Track $\mathcal{T} \in \mathcal{T}_{\text{bbox}}$ }
        \State $\mathcal{C}_{\text{track}}^{\mathcal{T}} = [F_1C_b, F_2C_b, \dots, , F_tC_b, \dots, F_TC_b] \ \forall \ b \in \mathcal{T}$ 
        \multicomment{Construct ``cluster track" $\mathcal{C}_{\text{track}}^{\mathcal{T}}$ of cluster labels $F_tC_b$ \\ at time $t$ for each bounding box $b$ in track $\mathcal{T}$.}
        \State $\mathcal{C}_{\text{track}} = \mathcal{C}_{\text{track}} \cup \mathcal{C}_{\text{track}}^{\mathcal{T}}$
        \Comment{Aggregate cluster tracks.}
    \EndFor

    \item[]
    \Statex \Comment{Algorithm continued on next page.}

\algstore{myalg}
\end{algorithmic}
\end{algorithm*}

\begin{algorithm*}                     
\begin{algorithmic}[1]                   
\algrestore{myalg}

    \item[] \Statex \textbf{Step 3c: Temporal Track Clustering}
    \State $\text{Jac} \leftarrow \text{JaccardScore}(\mathcal{C}_{\text{track}}^\mathcal{T}, \mathcal{C}_{\text{track}}^{\mathcal{T}'}) \ \forall \ \mathcal{T}, \mathcal{T}' \in \mathcal{T}_{\text{bbox}} $
    \Comment{Compute Jaccard Scores between pairs of tracks. }
    \State $\{\ell_{\mathcal{T}}, \mathcal{T} \}_{\ \forall \ \mathcal{T} \in \ \mathcal{T}_{\text{bbox}}} \leftarrow$ DBSCAN($\mathcal{T}_{\text{bbox}}$, Jac) 
    \multicomment{Cluster tracks $\mathcal{T}$ in $\mathcal{T}_{\text{bbox}}$, assigning object label $\ell_{\mathcal{T}}$ \\ for each $\mathcal{T} \in \mathcal{T}_{\text{bbox}}$ using Jaccard score of labels $\mathcal{C}_{\text{track}}^{\mathcal{T}}$.}
    

    \item[] \Statex \textbf{Step 4: Assemble Dataset}
    \State $\mathcal{D} = \{ \} $
    \Comment{Initialize Empty Dataset}

    \For{Frame $F_t \in \mathcal{V}$}
        \State $\mathcal{M}_t = \{\}$ \Comment{Initialize aggregated masks for frame $F_t$}
    
        \For{Object label $\ell \in \text{Label set} \  \mathcal{L}$}

            \For{$\mathcal{T} \in \mathcal{T}_{\text{mask}}$}
                \If{$\ell_{\mathcal{T}} == \ell$}
                    \State $\mathcal{M}_t = \mathcal{M}_t \cup m_t \in \mathcal{T}_{\text{mask}}$ 
                    \Comment{Collect all track masks $m_t$ at frame $F_t$, with object label $\ell$.}
                \EndIf
            \EndFor
            \State $m_{t, \ell} = \textsc{Aggregate}(\mathcal{M}_t)$ 
            \Comment{Aggregate these masks $\mathcal{M}_t$ into a single mask $m_{t, \ell}$ for object $\ell$.}

    
            \State Compute bounding box $b_{t,\ell} = \textsc{BBoxFromMask}(m_{t,\ell})$
            \State Item $\leftarrow \{F_t,\; b_{t,\ell}, \ m_{t,\ell},\ \ell\}$
            \Comment{Construct tuple of Frame $F_t$, Bounding Box $b_{t,\ell}$, Mask $m_{t,\ell}$, and Object Label $\ell$.}
            \State $\mathcal{D} \leftarrow \mathcal{D} \cup \{\text{Item}\}$
            \Comment{Aggregate this tuple into dataset $\mathcal{D}$.}
        \EndFor
    \EndFor
    





    \State \Return $\mathcal{D}$

\end{algorithmic}
\end{algorithm*}

\begin{algorithm}[h]
\caption{MOD Training}
\label{alg:mod_training}
\begin{algorithmic}[1]
\Require Dataset $\mathcal{D} = \{(F_t, b_t, L_i)\}_{t=1}^T$ from SODC, where $F_t$ is a video frame, $b_t$ is a bounding box, and $L_i$ is the object label.
\Require Pretrained Faster R-CNN model $\mathcal{M}$ with ResNet50 backbone.
\Require Standard RCNN loss function $\mathcal{L}(\cdot,\cdot)$.

\item[] \Statex \textbf{Step 1: Initialize Detector}
\State Initialize model $\mathcal{M}$ with ImageNet-pretrained weights.
\Statex \hfill \Comment{Load pretrained Faster R-CNN.}

\State Define image augmentations $\mathcal{A}$ as per \cref{app:mod}

\item[] \Statex \textbf{Step 2: Prepare Data}
\State Split $\mathcal{D}$ into training set $\mathcal{D}_{train}$ and validation set $\mathcal{D}_{val}$.

\item[] \Statex \textbf{Step 3: Optimize Detector}
\For{each epoch}
    \For{each batch $(F, B, L) \in \mathcal{D}_{train}$}
        \State $\tilde{F} \leftarrow \mathcal{A}(F)$
        \Statex \hfill         \Comment{Apply data augmentations.}

        \State $(\hat{B}, \hat{P}) \leftarrow \mathcal{M}(\tilde{F})$
        \Statex \hfill \Comment{Predict bounding boxes and class probabilities.}

        \State $\mathcal{J} \leftarrow \mathcal{L}((B,L),(\hat{B},\hat{P}))$
        \Statex \hfill \multicomment{Compute detection loss \\ (classification + regression).}

        \State Backpropagate $\mathcal{J}$ and update parameters of $\mathcal{M}$.
        \Statex \hfill \Comment{Gradient-based optimization step.}
    \EndFor

    \State Evaluate $\mathcal{M}$ on $\mathcal{D}_{val}$.
    \Statex \hfill \Comment{Monitor validation performance.}
\EndFor

\item[] \Statex \textbf{Step 4: Return Trained Model}
\State \Return Fine-tuned MOD detector $\mathcal{M}$.

\end{algorithmic}
\end{algorithm}

%% file: 4-2-table2.tex
{\renewcommand{\arraystretch}{1.3}
\begin{table*}[]
\begin{center}
\caption{Quantative evaluation of SODC + MOD, compared against supervised MOD baseline on standard detection metrics. Higher is better.  Highest values are bolded. MOD values are identical to \cref{table:1}}
\label{table:supervisedmod}
\resizebox{\linewidth}{!}{
\begin{tabular}{@{}cccccccc@{}}
\toprule
\multirow{2}{*}{Datasets} & \multicolumn{2}{c}{Approach} & \multicolumn{5}{c}{Metrics} \\ \cmidrule(l){2-8} 
 & Model & Prompt & mAP\textsubscript{0.5-0.95} & mAR\textsubscript{1} & F1\textsubscript{\rm 0.5-0.95} & Precision\textsubscript{\rm0.5-0.95} & Recall\textsubscript{0.5-0.95} \\ \midrule

\multirow{2}{*}{\begin{tabular}[c]{@{}c@{}}Meccano  \\ Dataset \cite{ragusa_MECCANO_2023} \end{tabular}}

 & MOD \textbf{(Ours)} & -- & \textbf{0.06} & \textbf{0.10} & \textbf{0.18} & \textbf{0.71} & \textbf{0.12} \\
 & Supervised MOD Baseline \textbf{(Ours)} & -- & 0.04 & 0.08 & 0.15 & 0.58 & 0.10 \\ \midrule

\multirow{2}{*}{\begin{tabular}[c]{@{}c@{}}In-House \\  Dataset \#1 \end{tabular}} 
 & MOD \textbf{(Ours)} & -- & 0.10 & 0.17 & \textbf{0.92} & \textbf{1.00} & \textbf{0.87} \\ 
 & Supervised MOD Baseline \textbf{(Ours)} & -- & \textbf{0.11} & \textbf{0.19} & 0.90 & \textbf{1.00} & 0.85 \\ 

 \bottomrule

\end{tabular}
}
\end{center}

\end{table*}
}